\documentclass[review]{elsarticle}

\usepackage{lineno,hyperref}
\usepackage{amsmath}
\usepackage{graphicx}
\usepackage{amssymb}
\usepackage{tabularx}
\usepackage{array}
\usepackage{bm}
\usepackage{epsfig}
\usepackage{algorithmic}
\usepackage{balance}
\usepackage[lined,linesnumbered,ruled]{algorithm2e}
\usepackage{url}
\usepackage{xcolor}
\usepackage[footnotesize]{caption}
\usepackage{multirow}
\usepackage{rotating}
\usepackage{lscape}
\usepackage{bm}
\usepackage{array}
\usepackage{url}
\usepackage {subfig,color}

\usepackage{amssymb} % define this before the line numbering.
\usepackage{multirow}
\usepackage{booktabs}
\usepackage{diagbox}

\RequirePackage{latexsym,amsmath,amssymb}

\modulolinenumbers[5]

\journal{Journal of Elsevier}

\begin{document}

\begin{frontmatter}

\title{SalFBNet: Learning Pseudo-Saliency Distribution via Feedback Convolutional Networks}

%\author{Guanqun Ding\fnref{myfootnote}}

\author[mymainaddress,mysecondaryaddress]{Guanqun Ding\corref{mycorrespondingauthor}}
\ead{guanqun.ding@aist.go.jp}

\author[mysecondaryaddress]{Nevrez \.{I}mamo\u{g}lu\corref{mycorrespondingauthor}}
\ead{nevrez.imamoglu@aist.go.jp}
\cortext[mycorrespondingauthor]{Corresponding author.}
\author[mysecondaryaddress]{Ali Caglayan} 
\author[mymainaddress,mysecondaryaddress]{Masahiro Murakawa}
\author[mysecondaryaddress]{Ryosuke Nakamura}

\address[mymainaddress]{Graduate School of Science and Technology, University of Tsukuba, Tsukuba, Japan}
\address[mysecondaryaddress]{National Institute of Advanced Industrial Science and Technology, Tokyo, 305-8567, Japan}

\begin{abstract}
Feed-forward only convolutional neural networks (CNNs) may ignore intrinsic relationships and potential benefits of feedback connections in vision tasks such as saliency detection, despite their significant representation capabilities. In this work, we propose a feedback-recursive convolutional framework (SalFBNet) for saliency detection. The proposed feedback model can learn abundant contextual representations by bridging a recursive pathway from higher-level feature blocks to low-level layers. Moreover, we create a large-scale Pseudo-Saliency dataset to alleviate the problem of data deficiency in saliency detection. We first use the proposed feedback model to learn saliency distribution from pseudo-ground-truth. Afterwards, we fine-tune the feedback model on existing eye-fixation datasets. Furthermore, we present a novel Selective Fixation and Non-Fixation Error (sFNE) loss to facilitate the proposed feedback model to better learn distinguishable eye-fixation-based features. Extensive experimental results show that our SalFBNet with fewer parameters achieves competitive results on the public saliency detection benchmarks, which demonstrate the effectiveness of proposed feedback model and Pseudo-Saliency data. 

% \textcolor{blue}{Feed-forward only convolutional neural networks (CNNs) may ignore the intrinsic relationship and potential benefits of the feedback connections in vision tasks such as saliency detection, despite their significant representation capabilities. In this work, we propose a feedback-recursive convolutional framework (SalFBNet) for saliency detection. The proposed feedback model can learn abundant contextual representations by bridging a recursive pathway from higher-level feature blocks to low-level layer. 
% Moreover, before fine-tuning the proposed framework on small-scale public saliency datasets, we explore the impact of large amount of pseudo annotations by initially training our feedback-recursive CNN model with fake ground-truth samples (Pseuso-Saliency labels) obtained from aggregating pseudo-annotators (\emph{i.e.} state-of-the-art saliency models) in a human-free manner. 
% Finally, in order to make the proposed framework better learn distinguishable eye-fixation-based features, we propose a novel Selective Fixation and Non-Fixation Error (sFNE) loss motivated by the Normalized Saliency Scanpath (NSS) metric to fine-tune the proposed model on public eye-fixation datasets.
% Extensive experimental results show that our SalFBNet with relatively fewer parameters achieves competitive results on the public saliency detection benchmarks, which demonstrate the effectiveness of initial pre-training with the pseudo-saliency dataset and fine-tuning with proposed sFNE loss.
% }
\end{abstract}


\end{frontmatter}

% \linenumbers

\section{Introduction}
\label{sec:intro}
\begin{figure}[t!]
    \centering
    \includegraphics[width =\textwidth]{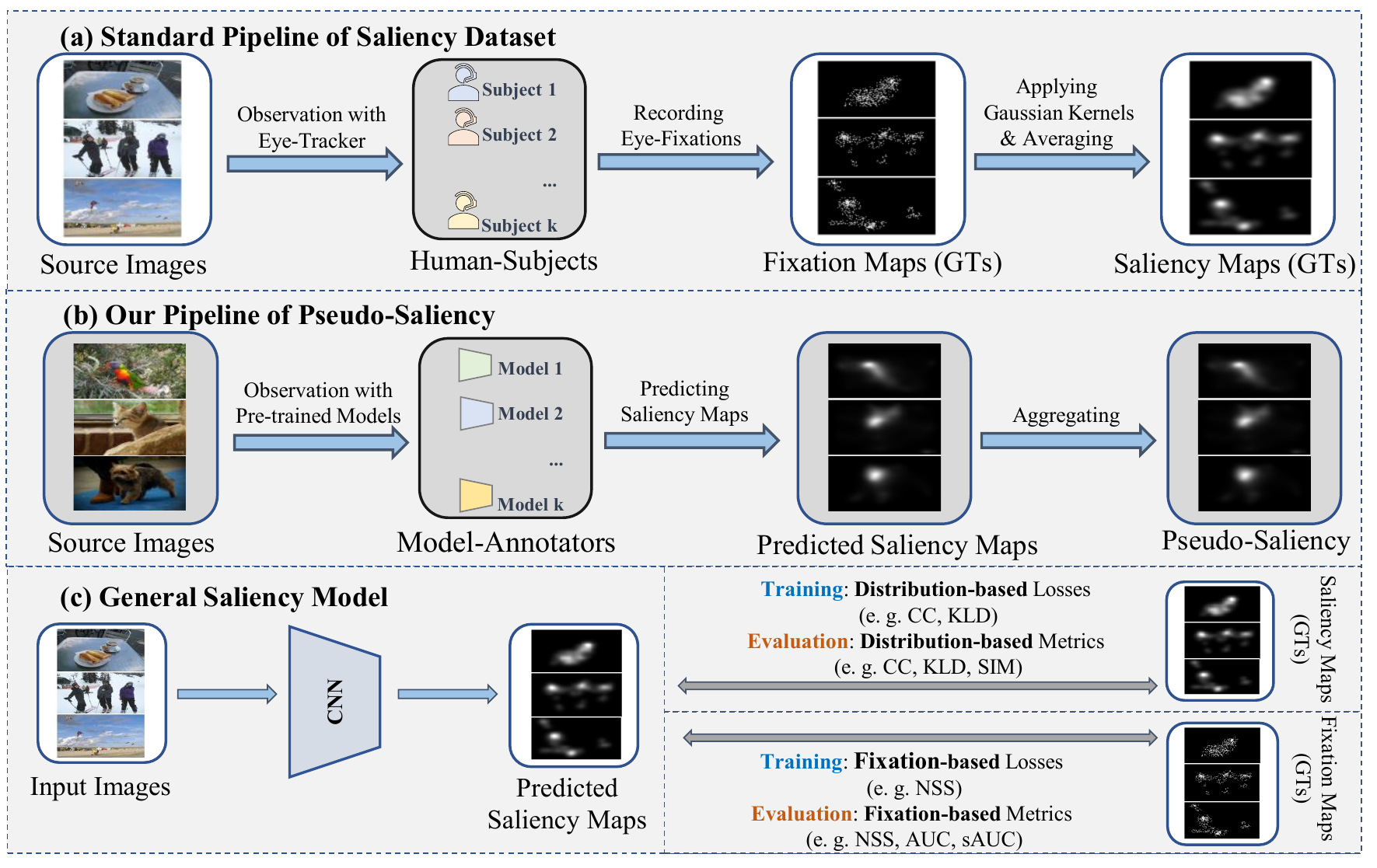}
    \caption{Pipelines of saliency annotation and general training/testing of saliency model. (a) Standard pipeline of eye-fixation dataset; (b) Our pipeline of pseudo saliency; (c) General training and testing of saliency models. 
    }
\label{fig:standard_saliency_data}
\end{figure}

Attention mechanism plays an important role in human visual system (HVS) by automatically focusing on the most relevant regions of observed scenes \cite{itti1998model}. %This cognitive selective mechanism allows us to quickly capture key information that results in a more precise interpretation of complex visual scenes. 
To better understand the attention mechanism of HVS, researchers often study human eye movement by recording a subject’s gaze while observing a stimuli (\emph{e.g.} an image) \cite{che2019gaze}, as shown in Figure \ref{fig:standard_saliency_data} (a). In this way, the fixations collected from eye-tracker indicate the most attractive locations of a scene. Typically, saliency dense map is generated from fixation maps with a Gaussian kernel to further represent the salient locations. This pixel-wise dense map denotes the attentive regions of the stimuli \cite{che2019gaze} (see Figure \ref{fig:standard_saliency_data} (a)). Saliency detection aims to find out the most informative and conspicuous fixations from a visual scene by simulating the similar attention mechanism of human eyes \cite{wang2017deep}. Saliency prediction on images/videos has been explored extensively in the past decades \cite{che2019gaze}. Fixation maps (fixation-based) collected from subjects and saliency maps (distribution-based) generated by the fixation maps are both regarded as the ground-truths (GTs) and used in the training and evaluation of saliency detection models \cite{che2019gaze, wang2017deep}, as illustrated in Figure \ref{fig:standard_saliency_data} (c). 

Saliency detection has been successfully applied to various computer vision tasks, such as compression \cite{fosco2020much} and image cropping \cite{chen2016automatic}. Although great progress has been made in the field, there are still various challenges exist to be investigated and addressed. First, most existing CNN-based saliency models mainly utilize forward-only pathway to learn visual representations \cite{wang2017deep, cornia2018predicting}, which ignores top-down connections of contextual features. In addition, some CNN-based saliency methods \cite{pan2016shallow} employ large model size with high computational cost to improve representation learning capability. Moreover, the lack of manually-labeled annotations may hinder further performance boosts of existing models. However, the collection and annotation of human gaze from eye-tracker task is extremely time-consuming and labor-intensive. 

To address the problem of insufficient training data for saliency prediction, we propose a similar pipeline of \emph{pseudo-saliency} labelling motivated by the standard subjective experiments \cite{che2019gaze, judd2012benchmark} of human gaze collection from eye-tracker, as demonstrated in Figure \ref{fig:standard_saliency_data} (b).
In our pipeline, we use pre-trained CNN models to \emph{annotate} saliency distribution of new RGB images. The well-known knowledge distillation (KD) method usually adopts teacher-student strategy to transfer softened knowledge from a large teacher network to a simple student model \cite{kim2021self}. 
%Generally, classification tasks utilize one-hot labels as supervision, however, KD approaches aim to provide auxiliary continuous class distribution/similarity during feature learning \cite{kim2021self}. 
Inspired by knowledge distillation models, we believe that the saliency distribution knowledge learned from these pre-trained models can be transferred into \emph{pseudo-saliency} annotations of new scenes. In this way, we can freely annotate saliency distribution of RGB images. These images and corresponding pseudo-annotations can be used for the training of saliency models. 
%Consequently, it is highly desirable to present a lightweight yet efficient saliency model to alleviate the gap between bottom-up and top-down contextual features from the observed visual stimuli.

\begin{figure}
    \centering
    \includegraphics[width = 0.86\textwidth]{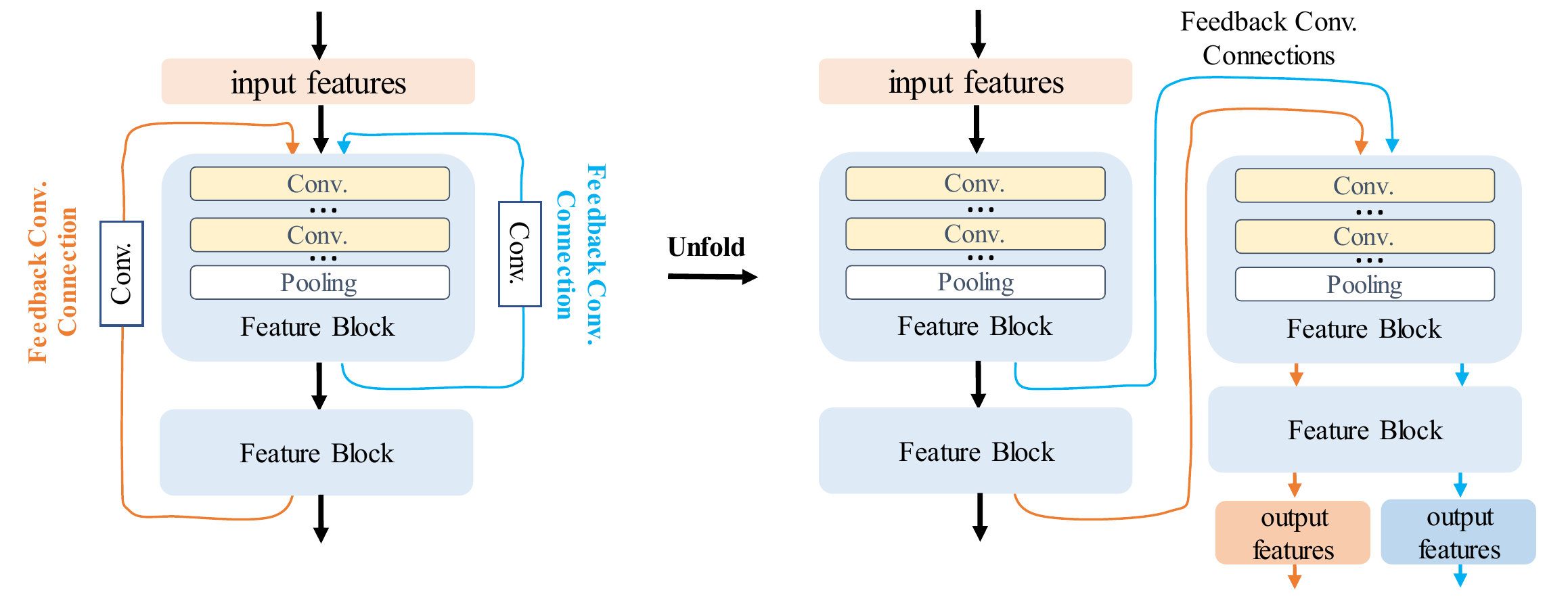}
    \caption{Simplified demonstration of feedback convolutional connections with two feature blocks. 
    }
\label{fig:fb_block}
\end{figure}

In biology, feedback mechanism is usually used to maintain the balance of the system by amplifying or suppressing the feedback signal (\emph{e.g.} positive or negative signal) \cite{cosentino2019feedback}. Likewise, human brain and visual system also leverage feedback mechanisms to process complex cognitive behaviors \cite{zamir2017feedback}. Feedback convolutional networks have been introduced to learn abundant representations to mimic the feedback mechanism in various computer vision tasks \cite{zamir2017feedback, cao2015look}. In our previous work \cite{ding2021fbnet}, we explore an extremely lightweight feedback recursive model by linking the feature pathway from high-level blocks to low-level layer for saliency detection. 
%The number of parameters of this model is only 1.18 M \cite{ding2021fbnet}. We demonstrate that utilizing feedback connections in this way is promising to improve the performance of feed-forward-only CNN model \cite{ding2021fbnet}. 
 
In this paper, we propose various model-based and experimental extensions to our previous work \cite{ding2021fbnet}. The contributions can be summarized as:
\begin{itemize}
    \item First, we extend our previous work \cite{ding2021fbnet} in various ways: i) To be a more general framework, the encoder of the proposed model can be flexibly substituted by different popular backbones, such as ResNet \cite{he2016deep}, DenseNet \cite{huang2017densely}, \emph{etc}. ii) To achieve that, we adopt feedback convolutional connections to bridge the pathway of top-down representations, as shown in Figure \ref{fig:fb_block}. 
    %This helps to better adapt the features from high-level blocks to low-level layers of any encoder used. 
    iii) Further, we utilize a new decoder with smoothing module as in \cite{droste2020unified} to incorporate the informative saliency scores from both forward- and feedback-streams. 
    \item Second, we investigate the use of pseudo annotations for the training of proposed feedback framework. We found that the feedback models can learn rich representations from human-free annotated data, which demonstrates that the knowledge of saliency distribution can be transferred by model-annotators. Also, we compare the performance boost of different initialization ways for saliency prediction.
    \item Furthermore, we propose a novel Selective Fixation and Non-Fixation Error (sFNE) loss to make the proposed model better at learning eye-fixation-based features. Our sFNE loss not only considers the cost at fixation points, but also calculates the error at randomly selected non-fixation points. 
\end{itemize}
Extensive experiments show the effectiveness of the proposed feedback framework and sFNE loss. Besides, we show that the feedback models can learn the information of saliency distribution from \emph{pseudo-saliency} annotations.

\section{Related Works}
\label{sec:sec_related}
\subsection{Saliency Detection on RGB Images}\label{sec_related_rgb}
Saliency detection research is mainly classified into human eye fixation prediction (\emph{i.e.} saliency map prediction) \cite{che2019gaze} and salient object detection (SOD) \cite{zhang2020learning, zhang2019synthesizing}. Eye fixation prediction focuses on predicting human gazes on images where human eyes are most attracted to \cite{che2019gaze}. On the other hand, SOD aims to identify the regions of salient objects from an image \cite{zhang2020learning, zhang2019synthesizing}. In this work, we focus on the prediction of human gaze from visual stimulus.

% Driven by the attention theory of HVS, saliency detection approaches can also be roughly divided into bottom-up model \cite{itti1998model} and top-down model \cite{qiu2020simple}. In the early ages, traditional saliency detection methods are usually developed based on bottom-up manner \cite{itti1998model}. Their assumption is that the salient fixations with distinctive features could be more attentive by comparing with the surrounding non-salient locations \cite{itti1998model}. 
% Thus, bottom-up modelling methods usually explore low-level visual features for saliency prediction, such as contrast/center priors or color features \cite{itti1998model}. 
% However, top-down models \cite{qiu2020simple} always utilize high-level distinctive and abstract representations to achieve higher performance \cite{qiu2020simple}. These studies show that the features from both bottom-up or top-down manner can effectively improve the performance of saliency detection models. 

In the early ages, traditional saliency detection methods are usually developed based on bottom-up manner \cite{itti1998model}. Recently, deep-learning based saliency methods \cite{cornia2018predicting, linardos2021deepgaze, droste2020unified} have achieved high performance through the superior capability of CNNs in feature representation. For example, Pan \emph{et al.} introduce a generative adversarial model for visual saliency prediction \cite{pan2017salgan}. Cornia \emph{et al.} develop a CNN model to combine the multi-layer features for saliency detection \cite{cornia2016multi}. 
%Furthermore, they propose a convolutional LSTM-based eye-fixation detection model to refine the predicted saliency map and learn the Gaussian prior map \cite{cornia2018predicting}. 
Kummerer \emph{et al.} propose a saliency detection model to explore the influence of low-level features (local intensity and contrast) and high-level features for eye-fixation prediction \cite{kummerer2017understanding}. Droste \emph{et al.} propose an unified saliency model based on domain adaptive for static images and dynamic videos \cite{droste2020unified}. Kroner \emph{et al.} use a Atrous Spatial Pyramid Pooling (ASPP) module to capture multi-scale convolutional features for saliency detection \cite{kroner2020contextual}. 
%Wang \emph{et al.} propose a densely supervised model to predict human eye fixation \cite{wang2017deep}. 
Jia \emph{et al.} establish an encoder-decoder-style network with expandable multi-layer for saliency prediction \cite{jia2020eml}. Fan \emph{et al.} utilize a context adaptive saliency network to learn the spatial and semantic context for gaze prediction \cite{fan2018emotional}. These models leverage a feed-forward-only architecture for saliency prediction, which may neglect the significant benefits of top-down features from feedback connection. In this work, we adopt feedback convolutional connections to bridge the feature pathway from high-level blocks to low-level layers to improve the representation ability.

\subsection{Pseudo Labelling for Saliency Detection}\label{sec_related_pseudo}

The problem of insufficient training data in saliency detection urges the emergence of several previous works to reduce the dependency of human annotated ground-truths \cite{zhang2020learning, zhang2019synthesizing, che2019gaze}. In the field of salient object detection (SOD), Nguyen \emph{et al.} propose a self-supervised CNN model with conditional random field (CRF) to refine the noisy pseudo-labels generated from several handcrafted methods \cite{nguyen2019deepusps}. Different from the study \cite{nguyen2019deepusps}, Zhang \emph{et al.} propose a joint learning strategy for salient object detection and noise modelling \cite{zhang2020learning}. Their hypothesis is that the pseudo-label of the input image can be represented as 
a combination of predicted saliency map and the noise map \cite{zhang2020learning}. In the study \cite{zhang2019synthesizing}, Zhang \emph{et al.} propose a deep saliency model by learning the synthesized weak-supervision generated by unsupervised salient object detectors.
It is mentioned in the study \cite{cheng2021highly} that the backbone initialized by pre-trained weights on ImageNet \cite{krizhevsky2012imagenet} is not necessary for salient object detection (SOD). The authors state that saliency models require far fewer parameters than the classification approach \cite{cheng2021highly}. In this work, we have reported similar findings in the field of eye fixation prediction.

However, to the best of our knowledge, there are a few works to address the problem of insufficient eye-fixation data due to the time-consuming gaze collection. Che \emph{et al.} explore the influence of different distortions on gaze prediction \cite{che2019gaze}. They adopt several transformations, such as cropping/rotation, to augment existing saliency datasets (\emph{e.g.} SALICON \cite{jiang2015salicon}). Then they collect eye-fixations of distorted image from eye tracker  \cite{che2019gaze}. 
% They found that the attentive fixations would be shifted severely by certain distortions such as Cropping; while some transformations have slight impact on human gaze such as Noise \cite{che2019gaze}.
Although this method \cite{che2019gaze} augments SALICON \cite{jiang2015salicon}, they still need to collect eye fixations from labor-intensive annotations. Besides, their method \cite{che2019gaze} does not provide data diversity from new visual scenes. Motivated by the standard subjective experiment of gaze collection, we annotate pseudo-labels by using pre-trained saliency models in this work.

\subsection{Feedback Convolutional Networks}\label{sec_related_feedback}
In recent years, feedback convolutional networks have been explored to learn top-down features in various computer vision applications \cite{zamir2017feedback, cao2015look, li2019feedback, deng2021deep}. For example, Zamir \emph{et al.} propose a feedback network based on convolutional LSTM for object recognition task \cite{zamir2017feedback}. They demonstrate that the feedback architecture can learn considerably different representations compared to the feed-forward counterpart \cite{zamir2017feedback}. Cao \emph{et al.} establish a feedback convolutional model to learn top-down attentive visual features for image classification and recognition \cite{cao2015look}. Stollenga \emph{et al.} use feedback connections for deep selective attention networks to improve the performance of classification \cite{stollenga2014deep}. Li \emph{et al.} propose to refine the low-level representation with high-level information by using a feedback convolutional network to improve the performance of image super-resolution  \cite{li2019feedback}. Deng \emph{et al.} introduce a deep-coupled feedback network for image exposure and super-resolution \cite{deng2021deep}. They use a multi-task learning strategy to optimize their feedback model \cite{deng2021deep}. 
% Rai \emph{et al.} propose a semantic feedback learning model to improve the semantic segmentation task of images in hot weather scenes \cite{rai2020spatial}. 
In our previous work \cite{ding2021fbnet}, we propose a lightweight feedback recursive network for image saliency prediction\cite{ding2021fbnet}. The works of \cite{zamir2017feedback, cao2015look, stollenga2014deep} show that feedback convolutional network can learn rich representations for various vision tasks.

\section{Proposed Method}
\label{sec:hs_saliency}
%In this section, we describe proposed architecture for saliency detection in detail. We first elaborate the component details of proposed feedback CNN model for human fixation prediction. After that, we introduce the pipeline of pseudo-saliency knowledge extraction.

% \subsection{Proposed Feedback CNN Framework}
% \label{sec:proposed_feedback_model}
% As demonstrated in Figure~\ref{fig:salfbnet}, there are three key components in the proposed model: i) An encoder with shared convolutional weights, ii) a feedback module that feeds learned forward features from high-level blocks to low-level layer, and iii) a decoder with deeply-supervision to fuse informative saliency scores.

\begin{figure*}[!t]
    \centering
    \includegraphics[width=\textwidth]{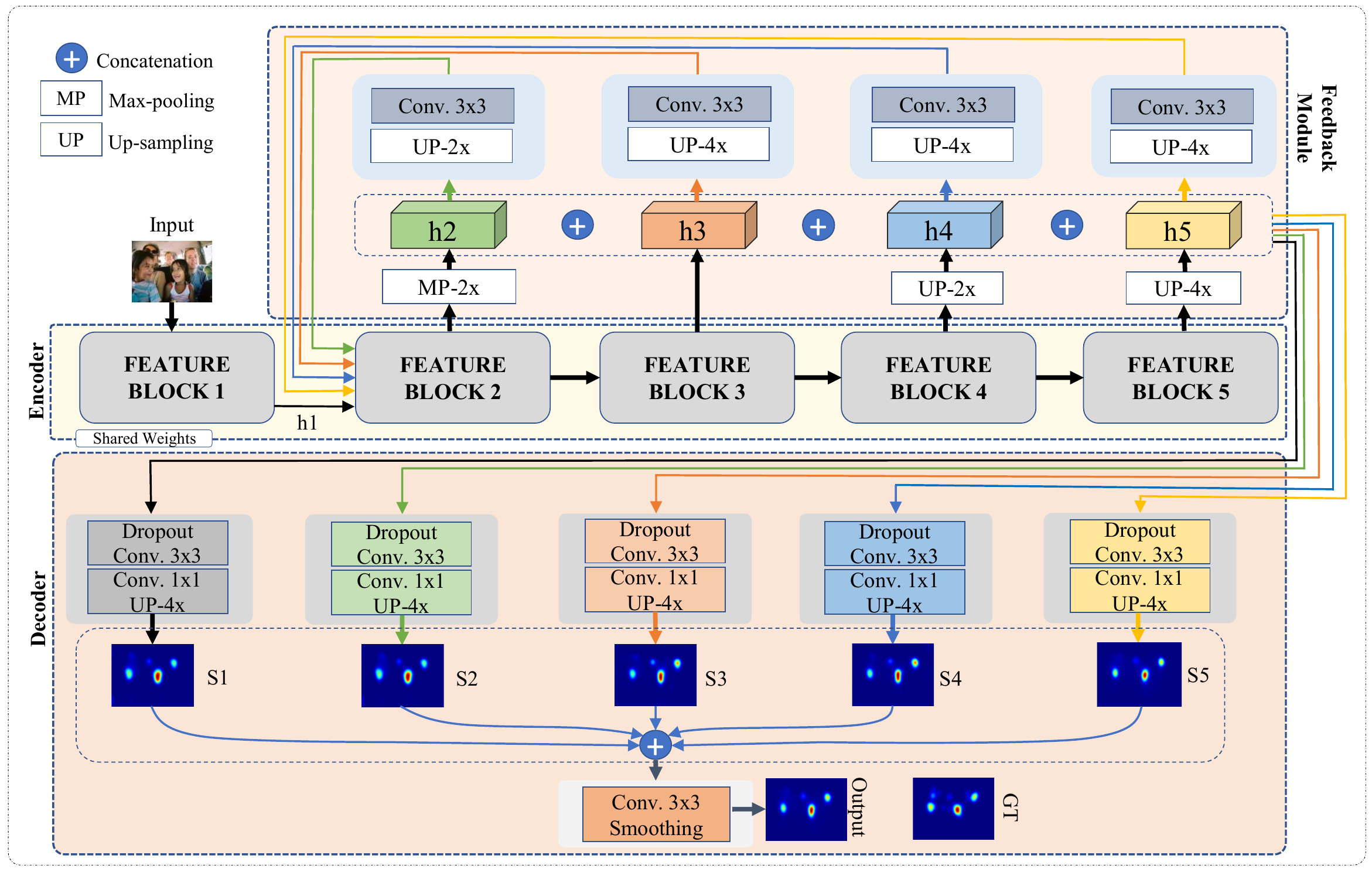}
    \caption{General architecture of proposed feedback convolutional network.}
\label{fig:salfbnet}
\end{figure*}

\subsection{Feed-Forward Feature Learning}
\label{sec:encoder}
First, we build a feed-forward feature extractor with five CNN blocks to learn multi-scale representations from input images, as depicted in Figure~\ref{fig:salfbnet}. The encoder of the proposed model can be flexibly substituted by popular CNN backbones, such as ResNet \cite{he2016deep}, VGG \cite{simonyan2015very}, or DenseNet \cite{huang2017densely} (see Section \ref{sec_ablation_backbones_pretraining}). 
%Alternatively, we can also use the lightweight feature extractor of our previous work \cite{ding2021fbnet} to make the feedback model more compact. 
%Different from ML-Net \cite{cornia2016multi} that combines the features \{\emph{h3}, \emph{h4}, \emph{h5}\} from the last three CNN blocks to obtain saliency prediction \cite{cornia2016multi}, we only neglect the first input feature block in this work. 
Afterwards, we fuse the multi-scale features \{\emph{h2}, \emph{h3}, \emph{h4}, \emph{h5}\} from all the other four CNN blocks. Then, the combined features are fed into the saliency decoder module to obtain the saliency score map (\emph{S1}) of the forward pass. Finally, we calculate the loss between \emph{S1} and eye-fixation ground-truth. 

\subsection{Feature Feedback Module}
\label{sec:feedback_module}
Next, we feed back the learned multi-scale features from high-level CNN blocks to low-level convolutional layers with shared CNN parameters. To this end, our intuitive objective is to enable the network to recursively aggregate contextual information into a holistic description through feedback connections. Since the first block includes the input layer of color images, we select the first layer of the second CNN block as the input layer of feedback convolutional features, as shown in Figure~\ref{fig:salfbnet}.

We first extract the forward-feature \emph{h2} from \emph{feature block 2}. After that, the \emph{h2} is fed back to the feed-forward feature extractor with a convolutional feedback connection and shared CNN weights. The feedback connection consists of an up-sampling layer and a convolutional layer with $3 \times 3$ kernel. Thus, we can obtain the multi-scale feedback features (see the green arrows in Figure~\ref{fig:salfbnet}). The output feedback features $\Hat{h}_{k}^{l}$ of $l$-th ($l \in \{2, 3, 4, 5\}$) CNN block from $k$-th ($k \in \{2, 3, 4, 5\}$) forward feature can be formulated as follows:
\begin{eqnarray}
    \Hat{h}_{k}^{l} = f_{k}^{l} ( \varphi (UP({h}_{k}) * {W}_{k}^{fb} ))
\label{equ:fb_feature}
\end{eqnarray}
where ${W}_{k}^{fb}$ is the convolutional weights of $k$-th feedback connection; $h_{k}$ is the feed-forward feature of $k$-th CNN block; $*$ denotes the convolutional operation; the $UP$, $\varphi$, and $f_{k}^{l}$ represent the up-sampling, activation function, and the $l$-th convolutional block of $k$-th forward feature, respectively. Similar with the forward saliency head of the decoder, the saliency score \emph{S2} is predicted with feedback saliency head by using the feedback features $\Hat{h}_{k}^{2}$. 

Following the process in utilizing the forward feature \emph{h2}, the forward feature \emph{h3} (orange arrows) from \emph{block 3}, the forward feature \emph{h4} (blue arrows) from \emph{block 4}, and the forward feature \emph{h5} (yellow arrows) from \emph{block 5} are also recursively fed back to the feed-forward feature extractor using feedback connections and the shared CNN weights, according to Equation \ref{equ:fb_feature}. Hence, we obtain the enhanced informative feedback features $\Hat{h}_{k}^{l}$ for each feed-forward feature ${h}_{k}$. Finally, the learned feedback features $\Hat{h}_{k}^{l}$ are concatenated and passed to saliency heads of decoder for predicting the saliency scores \emph{S3}, \emph{S4}, and \emph{S5}. 

\subsection{Saliency Feature Decoder}
\label{sec:decoder}
In the proposed feedback model, we utilize a feature decoder to aggregate the saliency scores from both forward and feedback pathways, as shown in Figure~\ref{fig:salfbnet}. The concatenated multi-scale features are used to predict the saliency score with individual saliency head in the decoder. The saliency scores $\{S_{n}|n \in \{1, 2, 3, 4, 5\} \}$ can be represented as:
\begin{eqnarray}
    S_{n} = UP ( \varphi ( \varphi (Dropout(H_{n}) * {W}_{n}^{3 \times 3} ) * {W}_{n}^{1 \times 1} ) ))
\label{equ:sal_score}
\end{eqnarray}
where $H_{n}$ denotes the fused features from encoder; ${W}_{n}^{3 \times 3}$ and ${W}_{n}^{1 \times 1}$ indicate the convolutional weights with $3 \times 3$ and $1 \times 1$ kernels for $n$-th saliency score, respectively; $Dropout$ is a dropout layer. Note that if $n=1$, the saliency score $S1$ is predicted by the forward pathway; otherwise, the saliency score is calculated by the feedback feature.

Next, we concatenate the informative saliency scores $S_{n}$ from a forward saliency head and four feedback saliency heads, as illustrated in Figure \ref{fig:salfbnet}. Furthermore, we predict a final saliency map $\Hat{\mathbf{S}}$ by using a fusion convolutional layer with $1 \times 1$ kernel and a smoothing convolutional layer with $41 \times 41$ kernel, which can be formulated as follows: 
\begin{eqnarray}
    \Hat{\mathbf{S}} = \varphi (Cat[S_{n}] * {W}^{fusion} ) * {W}^{smoothing}
\label{equ:final_score}
\end{eqnarray}
where $n \in \{1, 2, 3, 4, 5\}$ denotes the indices of saliency scores; $Cat[~]$ is the concatenation operation; ${W}^{fusion}$ and ${W}^{smoothing}$ represent the convolutional weights of fusion and smoothing layer, respectively.

\subsection{Loss Function}
\label{sec:loss_function}
As stated in \cite{kummerer2018saliency}, normalized scanpath saliency (NSS) metric is often used to measure performance of a saliency model. NSS is generally calculated by averaging normalized saliency values at eye-fixation pixels. Typically, the higher the NSS value, the better the performance of a model. Thus, there are several works \cite{fosco2020much, droste2020unified} adopt negative NSS (\emph{i.e.} -NSS) as loss function to train saliency models. However, the loss function of -NSS only considers the cost of normalized saliency at salient fixations, while ignoring the loss at non-salient fixations. 

\begin{figure}[!]
    \centering
    \includegraphics[width=0.86\textwidth]{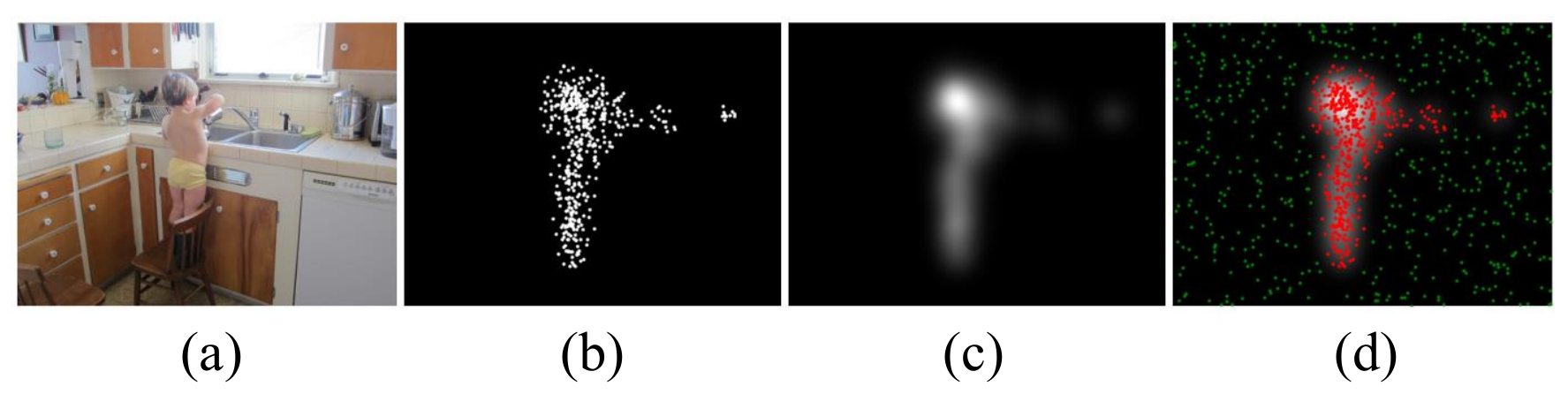}
    \caption{Illustration of Selective Fixation and Non-Fixation Error (sFNE) loss. From left to right: (a) image, (b) ground-truth fixation map, (c) ground-truth saliency map, (d) visualization of fixations (red points) in (b) and randomly-selected non-fixations (green points).}
\label{fig:selective_loss}
\end{figure}

Inspired by NSS metric and variant NSS loss in the studies \cite{jia2020eml, droste2020unified}, we introduce a novel fixation-based loss referred as Selective Fixation and Non-Fixation Error (sFNE) for training proposed feedback saliency model. We first randomly select the same number of non-fixations based on the fixations in the ground-truth map to ensure the balance between positive and negative samples in a normalized saliency map, as demonstrated in Figure~\ref{fig:selective_loss}. Our key idea is that the distribution between normalized prediction and ground-truth saliency map at both fixation and non-fixation locations should be as close as possible. To this end, we use sFNE to measure the difference of saliency distribution. Let $P, G, F, \bar{F}$ denote predicted saliency map, ground-truth saliency map, ground-truth fixation map, and selected non-fixation map, respectively. 
After normalization, we obtain normalized prediction $P'$ and normalized label $G'$.
% First, the normalized prediction $P'$ and normalized label $G'$ can be calculated as:
% \begin{eqnarray}
%     P'=\frac{P-\mu(P)}{\sigma(P)} \\
%     G'=\frac{G-\mu(G)}{\sigma(G)}
% \label{equ:sfe_loss_normalize}
% \end{eqnarray}
% where $\mu$ and $\sigma$ represent the mean and standard variance of the saliency map, respectively. 
Then, the loss of fixation and non-fixation locations can be computed as:
\begin{eqnarray}
    \mathcal{L}_{fixation} =\frac{1}{N} \sum_{i=1}^{N} \| P'_{i} \times F_{i} - G'_{i} \times F_{i} \|^{2} \\
    \mathcal{L}_{nonfixation} =
    \frac{1}{N} \sum_{j=1}^{N} \| P'_{j} \times \bar{F}_{j} - G'_{j} \times \bar{F}_{j} \|^{2}
\label{equ:sfe_loss_fix_nonfix}
\end{eqnarray}
where $F_{i} \in \{0, 1\}$ and $\bar{F}_{j} \in \{0, 1\}$; $N$ denotes the number of fixations, which can be written as: 
\begin{eqnarray}
    N=\sum_{i}F_{i}=\sum_{j}\bar{F}_{j}
\label{equ:sfe_loss_number}
\end{eqnarray}
where $i, j$ represent the indices of fixation and non-fixations, respectively. Therefore, sFNE loss $\mathcal{L}_{sFNE}$ can be formulated as follows:
\begin{eqnarray}
    \mathcal{L}_{sFNE} =\alpha \mathcal{L}_{fixation} + \beta \mathcal{L}_{nonfixation}
\label{equ:sfe_loss}
\end{eqnarray}
where $\alpha, \beta$ are the weights of the fixation and non-fixation costs, respectively. 

In addition to the fixation-based sFNE loss, we also combine other popular distribution-based loss \cite{yang2019dilated, fosco2020much} in the field of saliency detection. In this work, we combine our sFNE loss with Correlation Coefficient (CC) loss and Kullback-Leibler Divergence (KLD) loss (see \cite{bylinskii2018different, jia2020eml} for formulations). Therefore, the combined loss of proposed feedback model can be represented as follows:
\begin{eqnarray}
    \mathcal{L} =\gamma \mathcal{L}_{KLD} + \delta \mathcal{L}_{CC}+ \eta \mathcal{L}_{sFNE}
\label{equ:combined_loss}
\end{eqnarray}
where $\gamma, \delta, \eta$ are the weighting constants of the three losses, respectively. Note that we utilize the $\mathcal{L}_{CC}=1-CC$ as CC loss to avoid the negative value.

Besides, we propose to deeply supervise the saliency scores from forward and feedback pathways with ground-truths. In this way, the proposed model can learn abundant enhanced multi-scale features for saliency prediction. More specifically, the total loss of the $n$ predicted saliency scores $\{S_{n}, n \in \{1,2,3,4,5\}\}$ (\emph{i.e.} saliency maps of the forward saliency head and four feedback saliency heads) can be represented as follows:
\begin{eqnarray}
    Loss_{score} = \frac{1}{N} \sum \mathcal{L} (S_{n}, G, F, \bar{F})
\label{equ:loss1}
\end{eqnarray}
where $n$ and $N$ are the index and number of the predicted saliency score, respectively. Finally, we measure the cost between the final fused saliency prediction $\Hat{\mathbf{S}}$ and ground truth map by the following function: 
\begin{eqnarray}
    Loss_{fuse} = \mathcal{L} (\Hat{\mathbf{S}}, G, F, \bar{F})
\label{equ:lossf}
\end{eqnarray}

Thus, the overall loss for optimizing the proposed feedback saliency model can be calculated as:
\begin{eqnarray}
    Loss = \lambda_{1} Loss_{score} + \lambda_{2} Loss_{fuse}
\label{equ:lossall}
\end{eqnarray}
where the hyper-parameters $\lambda_{1}, \lambda_{2}$ are used for weighting the losses.

\subsection{Saliency Knowledge Transferring}
\label{sec:knowledge-transfer}

Motivated by the teacher-student strategy, we postulate that existing state-of-the-art pre-trained saliency models may have better initial knowledge of saliency distribution.
%instead of transfer learning based on CNN parameters learned from other vision tasks (e.g. image classification). 
Therefore, we conducted experiments to see if they can serve as better transmitters for transferring saliency knowledge to a new simple student model. To achieve this, starting with a set of images $\mathcal{X}=\{x_{i}, i=1,...,N\}$, where $N$ is the number of images, we select arguably top five models from the benchmark lists of state-of-of-the-art methods in public databases\footnote{https://saliency.tuebingen.ai/results.html}. Then we use these $M$=5 pre-trained models to \emph{annotate} the images $\mathcal{X}$. Similar with the standard subjective experiment of eye-fixation collection, we leverage these $M$ pre-trained models to generate the saliency probability maps $\mathcal{S}$ = $\{s_{i}^{j}, i=1,...,N; j=1,...,M\}$, where $s_{i}^{j}$ denotes the predicted probability map of $i$-th image with $j$-th pre-trained model-annotator. In standard human gaze experiment, the ground-truth saliency map is calculated by recorded fixations with a Gaussian kernel to represent the saliency probability. However, since the pre-trained saliency model naturally predicts saliency probability of an input image, we directly aggregate the predictions of the pre-trained model-annotators to avoid the biases from various models. Specifically, the aggregated saliency map can be formulated as follows:
\begin{eqnarray}
    {\Bar{s}}_{i}= g (\sum_{j=1}^{M} \alpha {s}_{i}^{j})
\label{equ:average}
\end{eqnarray}
where $\Bar{s}_{i}$ denotes the aggregated prediction of the results from $M$ pre-trained models; $\alpha$ represents the weight for weighting the predicted distribution; $g$ is a normalization function. Here, $\alpha$=0.2 in this work. Thus, the assembled probability maps can be represented as $\mathcal{\Bar{S}}=\{\Bar{s}_{i}, i=1,...,N\}$.

\begin{figure}
    \centering
    \includegraphics[width=0.86\textwidth]{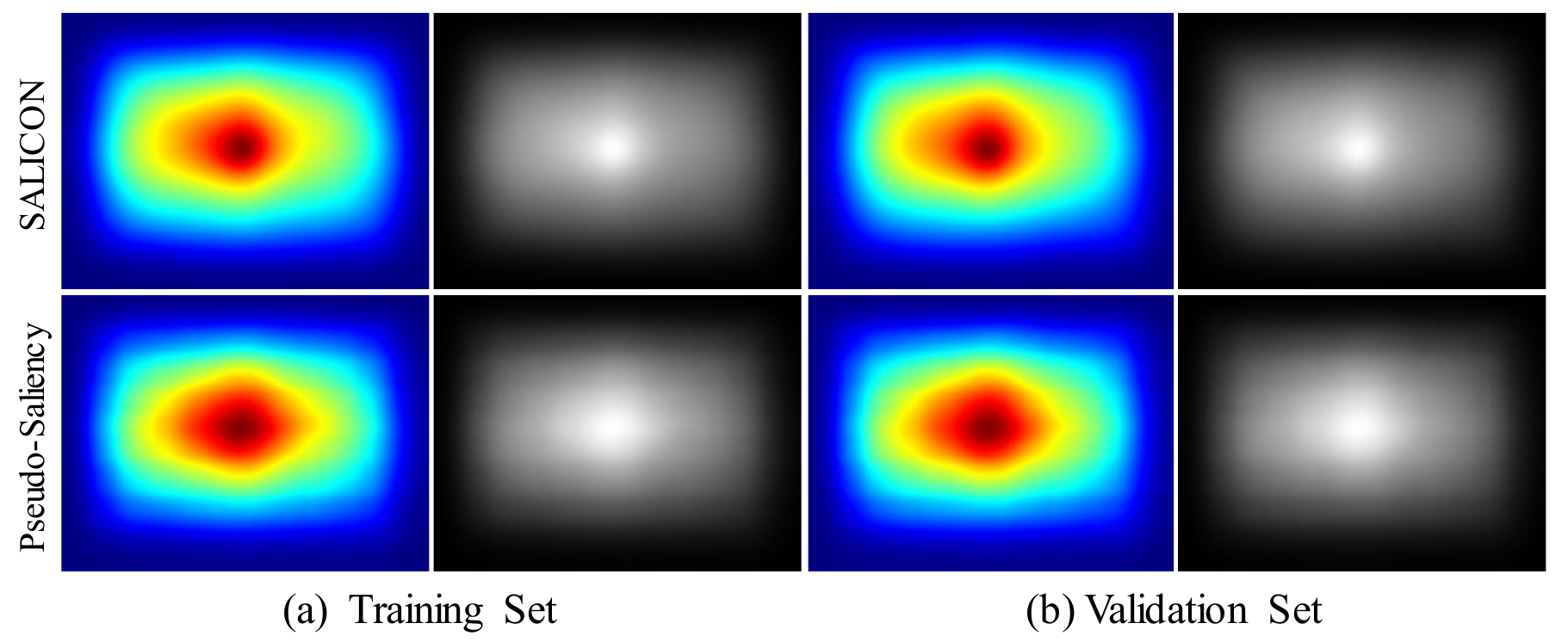}
    \caption{Illustration of saliency distribution. The top and bottom are averaged saliency distribution from SALICON \cite{jiang2015salicon} and our Pseudo-Saliency dataset, respectively.}
\label{fig:mean_sal}
\end{figure}

To validate the aggregated pseudo-saliency probability map, we visualize the averaged saliency distribution and corresponding heat map on SALICON \cite{jiang2015salicon} in Figure \ref{fig:mean_sal}. We observe that the pseudo-saliency distribution approximate to the distribution of human annotations. This validates our hypothesis that the saliency knowledge of the pre-trained models can be transferred to aggregated pseudo-labels. In this way, we can \emph{annotate} the pseudo saliency of RGB image from any published datasets in a human-free manner. 
%Generally, existing saliency models \cite{cornia2016multi, yang2019dilated} use the backbone with pre-trained weights on ImageNet \cite{krizhevsky2012imagenet} as their feature encoders. This is because of the lack of human-labelled training samples in saliency prediction (\emph{e.g.} 10k training samples in SALICON \cite{jiang2015salicon}). Clearly, this might limit the use of a new lightweight feature extractor. 
As proposed in this work, Pseudo-Saliency data can be an alternative way to train a new CNN model from scratch. 

%In order to increase the scene diversity of Pseudo-Saliency dataset, we select images from ImageNet \cite{krizhevsky2012imagenet} and several public SOD datasets. Then we create a new large-scale Pseudo-Saliency dataset\footnote{We will release codes and Pseudo-Saliency data at \url{https://github.com/gqding/SalFBNet}.} with more than 176k \emph{model-annotated} samples. We believe that the aggregated saliency knowledge of pre-trained models behind in pseudo-ground-truths can be transferred to new CNN networks (see Section \ref{sec_ablation_backbones_pretraining}). 

\section{Experimental Results}
\label{sec:experiments}
%In this section, we first give a comprehensive discussion of model training on Pseudo-Saliency dataset. Next, we show experimental results and analysis of proposed feedback architecture with different backbones. Finally, we compare proposed feedback model with existing saliency prediction methods. 

\subsection{Experimental Setup}\label{sec_experiment}

% \subsubsection{Pseudo-Saliency Annotation}\label{pseudo-saliency}

% \subsubsection{Evaluation Dataset}\label{dataset}
In this work, we utilize six commonly-used saliency datasets, including SALICON \cite {jiang2015salicon}, MIT1003 \cite{judd2009learning}, MIT300 \cite{judd2012benchmark}, DUT-OMRON \cite{yang2013saliency}, PASCAL-S \cite{li2014secrets}, and TORONTO \cite{bruce2007attention}. We show the detailed information of these datasets in Table~\ref{tab:comp_datasets}. The large-scale SALICON \cite {jiang2015salicon} dataset officially contains training set, validation set, and testing set for saliency prediction. 
%MIT1003 \cite{judd2009learning} totally consists of 1003 color images and corresponding eye-fixation maps recorded by eye-tracker. 
We randomly divide MIT1003 \cite{judd2009learning} into 900 images and 103 images for training set and validation set, respectively. After pre-training on Pesudo-Saliency dataset, we fine-tune our feedback models on SALICON \cite {jiang2015salicon} and MIT1003 \cite{judd2009learning}. Other datasets are used for performance testing, as illustrated in Table~\ref{tab:comp_datasets}.

\begin{table}[!t]
\centering \caption{Details of different saliency datasets.}
\resizebox{0.86\textwidth}{!}{
\begin{tabular}{  c | c  c  c  c  c  }
\hline
\small Dataset & \#Image & \#Training & \#Val. & \#Testing &  Size   \\
\hline\hline
\small MIT300 \cite{judd2012benchmark} & 300  & - & - & 300  & $\sim$ 44.4 MB  \\
\small PASCAL-S \cite{li2014secrets} & 850  & - & - & 850  & $\sim$ 108.3 MB  \\
\small DUT-OMRON \cite{yang2013saliency} & 5,168  & - & - & 5,168  & $\sim$ 151.8 MB \\
\small TORONTO \cite{bruce2007attention} & 120  & - & - & 120  & $\sim$ 92.3 MB  \\
\small SALICON \cite{jiang2015salicon} & 20,000  & 10,000 & 5,000 & 5,000  & $\sim$ 4 GB  \\
\small MIT1003 \cite{judd2009learning} & 1003  & 900* & 103* & -  & $\sim$ 178.7 MB \\
\small Pseudo-Saliency (Ours) & 176,880  & 150,000 & 26,880 & -  & $\sim$ 24.2 GB  \\
\hline
\multicolumn{6}{l}{~\small *The training set and validation set are randomly split in this work.}  \\
\end{tabular}
}
\label{tab:comp_datasets}
\end{table}

% \subsubsection{Implementation Details}\label{implementation}
The proposed feedback models are implemented with PyTorch library. We utilize SGD optimizer for optimization. Batch size, momentum and weight decay values are 10, 0.9 and 1e-4, respectively. 
%For pre-training on Pseudo-Saliency dataset, initial learning rate is 4e-2. For fine-tuning on existing saliency dataset, initial learning rate is 1e-4. 
The initial learning rates of pre-training and fine-tuning are 4e-2 and 1e-4, respectively. 
The learning rate decay value is set as 0.9. In our model, factor of dropout is 0.5 and activation function $\varphi$ is ReLU. In Equation (\ref{equ:sfe_loss}) and (\ref{equ:lossall}), the weights of $\alpha, \beta, \lambda_{1}, \lambda_{2}$ are set as 1. In Equation (\ref{equ:combined_loss}), we use $\gamma$=1, $\delta$=0.1, and $\eta$=0.025, respectively. 

For pseudo-saliency annotation, we first select 150,000 color images from widely-used ImageNet \cite{krizhevsky2012imagenet} dataset and 26,880 color images from SOD datasets. In our experiment, SOD datasets include CSSD \cite{yan2013hierarchical}, ECSSD \cite{shi2015hierarchical}, HKU-IS\cite{li2015visual}, MSRA-B\cite{jiang2013salient}, MSRA10K\cite{cheng2014global}, and THUR15K\cite{cheng2014salientshape}. Afterwards, we choose M=5 pre-trained saliency models to \emph{annotate} these images. The pre-trained models include DeepGazeIIE \cite{linardos2021deepgaze}, UNISAL \cite{droste2020unified}, MSINet \cite{kroner2020contextual}, EMLNet \cite{jia2020eml}, CASNetII \cite{fan2018emotional}. We directly use their pre-trained weights and default settings for inference of saliency distribution. Therefore, we create a large-scale Pseudo-Saliency dataset containing 176,880 color images and corresponding pseudo-ground-truths. Our Pseudo-Saliency dataset is divided into training set and validation set, as shown in Table~\ref{tab:comp_datasets}.

% \subsubsection{Evaluation Metrics}\label{Metrics}
Similar with the studies \cite{droste2020unified, kummerer2018saliency}, we use popular metrics to report performance results, including linear correlation coefficient (CC), area under ROC curve (AUC), shuffled AUC (sAUC), normalized scanpath saliency (NSS), similarity (SIM), information gain (IG) and Kullback-Leibler divergence (KLdiv). Note that the larger the value of IG, CC, AUC, sAUC, and NSS, and the smaller the value of KLdiv, the better the performance of the saliency model (see the study \cite{kummerer2018saliency} for more information about these metrics). For a fair comparison, we utilize same implementations of these metrics\footnote{https://github.com/cvzoya/saliency/tree/master/code\_forMetrics} for performance evaluation.

\subsection{Comparison with Existing Saliency Methods}
\label{sec_existing_methods}
In this experiment, we compare our feedback models with existing saliency prediction methods on 5 popular eye-fixation datasets, including SALICON \cite{jiang2015salicon}, MIT300 \cite{judd2012benchmark}, DUT-OMRON \cite{yang2013saliency}, PASCAL-S \cite{li2014secrets}, and TORONTO \cite{bruce2007attention}. 

\begin{table*}[!t]
\centering \caption{Quantitative comparison with 8 existing methods on SALICON \cite{jiang2015salicon} testing set. \textbf{Pub.} means the publication of the method. The best three results are respectively shown in \textcolor{red}{red}, \textcolor{blue}{blue}, and \textcolor{green}{green} color.}
\resizebox{0.94\textwidth}{!}{
\begin{tabular}{  c |  c | c  c  c  c  c  c  c }
\hline
\small Model & Pub. & \small  AUC $\uparrow$ & \small sAUC $\uparrow$ &  \small IG $\uparrow$   &  \small  NSS $\uparrow$   & \small  CC $\uparrow$  & \small  SIM $\uparrow$ & \small  KLdiv $\downarrow$ \\
\hline\hline
\small MD-SEM \cite{fosco2020much} & CVPR2020  & 0.8640  & \textcolor{red}{0.7460}  & 0.6600 & \textcolor{red}{2.0580} & 0.8680      & 0.7740 & 0.5680  \\
\small EMLNet \cite{jia2020eml} & IVC2020  & \textcolor{green}{0.8660}  & \textcolor{red}{0.7460}  & 0.7360 & \textcolor{blue}{2.0500} & 0.8860      & 0.7800 & 0.5200  \\
\small SAM-Res \cite{cornia2018predicting} & TIP2018   & 0.8650 &  \textcolor{blue}{0.7410}  & 0.5380 & \textcolor{green}{1.9900} & \textcolor{blue}{0.8990}     & \textcolor{blue}{0.7930} & 0.6100  \\
\small ACNet-V17 \cite{li2021attention} & NC2021  & \textcolor{green}{0.8660}    & 0.7390  & \textcolor{red}{0.8540} & 1.9480 & \textcolor{green}{0.8960}      & \textcolor{green}{0.7860} & \textcolor{red}{0.2280}  \\
\small DI-Net \cite{yang2019dilated} & TMM2019      & 0.8620    & 0.7390  & 0.1950 & 1.9590 & \textcolor{red}{0.9020}   & \textcolor{red}{0.7950} & 0.8640  \\
\small MSI-Net \cite{kroner2020contextual} & NN2020    & 0.8650    & 0.7360  & 0.7930 & 1.9310 & 0.8890   & 0.7840 & 0.3070  \\
\small GazeGAN \cite{che2019gaze} & TIP2019   & 0.8640   & 0.7360  & 0.7200 & 1.8990 & 0.8790    & 0.7730 & 0.3760  \\
\small FBNet \cite{ding2021fbnet} & MVA2021   & 0.8430   & 0.7060  & 0.3430 & 1.6870 & 0.7850    & 0.6940 & 0.7080  \\
\hline
\small \textbf{SalFBNet-Res18 (Ours)}  & -      & \textbf{\textcolor{blue}{0.8670}} & \textbf{0.7330}  & \textbf{\textcolor{green}{0.8050}} & \textbf{1.9500} & \textbf{0.8880}    & \textbf{0.7730} & \textbf{\textcolor{green}{0.3030}} \\
\small \textbf{SalFBNet-Res18Fixed (Ours)}  & -   & \textbf{\textcolor{red}{0.8680}}  & \textbf{\textcolor{green}{0.7400}}  & \textbf{\textcolor{blue}{0.8390}} & \textbf{1.9520} & \textbf{0.8920}   & \textbf{0.7720} & \textbf{\textcolor{blue}{0.2360}} \\
\hline
\multicolumn{9}{l}{~\small Note that all the metric values are collected from the SALICON \cite{jiang2015salicon} leaderboard and the literature.}  \\
\end{tabular}
}
\label{tab:salicon_testing}
\end{table*}

For SALICON \cite{jiang2015salicon} dataset, we evaluate the results of two feedback models: SalFBNet-Res18 and SalFBNet-Res18Fixed. The backbone of SalFBNet-Res18 is ResNet18 \cite{he2016deep}, while the backbone of SalFBNet-Res18Fixed is ResNet18 with fixed filter size of 128. After finetuning on SALICON \cite{jiang2015salicon}, we submit our prediction results of the testing set to the official server of SALICON \cite{jiang2015salicon} for evaluation. We report the quantitative results of SALICON \cite{jiang2015salicon} in Table \ref{tab:salicon_testing}, including 8 existing saliency methods. We can see that our two feedback models rank in the top three for the most metrics and achieve competitive performances compared with the existing methods. Especially, the AUC value of SalFBNet-Res18Fixed achieves the best. Additionally, SalFBNet-Res18Fixed outperforms SalFBNet-Res18 on most metrics. These comparison results further verify the finding that the feedback model with few number of parameters can also learn rich representation for saliency prediction. Also, these two extended feedback models outperform our previous FBNet \cite{ding2021fbnet} model by a large margin. These results demonstrated the effectiveness of the feedback framework and our training scheme with pseudo-saliency data and sFNE loss.

\begin{table*}[!t]
\centering \caption{Quantitative comparison with 24 existing methods on MIT300 \cite{judd2012benchmark} testing set. \textbf{D} and \textbf{T} means the CNN-based and traditional-based method. \textbf{Pub.} means the publication of the method. The best three results are respectively shown in \textcolor{red}{red}, \textcolor{blue}{blue}, and \textcolor{green}{green} color.}
\resizebox{0.86\textwidth}{!}{
\begin{tabular}{  c | c|  c |  c  c  c  c  c  c }
\hline
\small Model & Pub. & D/T &  \small AUC $\uparrow$   &  \small  sAUC $\uparrow$   & \small  NSS $\uparrow$ & \small  CC $\uparrow$ & \small  KLdiv $\downarrow$ & \small  SIM $\uparrow$ \\
\hline\hline
\small CovSal \cite{erdem2013visual}  & JoV2013 & T   & 0.8116 & 0.5894 & 1.3362    & 0.5000  & 1.7220 & 0.5058  \\
\small LDS \cite{fang2016learning}  & TNNLS2016 & T     & 0.8108 & 0.6020 & 1.3649    & 0.5177  & 1.0631 & 0.5222  \\
\small Judd \cite{judd2009learning}  & ICCV2009 & T     & 0.8095 & 0.6003 & 1.1882    & 0.4664  & 1.1084 & 0.4182  \\
\small GBVS \cite{harel2006graph}  & NIPS2006 & T     & 0.8062 & 0.6299 & 1.2457    & 0.4791  & 0.8878 & 0.4835  \\
% \small FES \cite{tavakoli2011fast}  & SCIA2011 & T     & 0.8018 & 0.5941 & 1.2763    & 0.4827  & 2.3018 & 0.4919  \\
\small BMS \cite{zhang2013saliency}  & ICCV2013 & T     & 0.7718 & 0.6918 & 1.1512    & 0.4130  & 1.0235 & 0.4456  \\
% \small RARE \cite{riche2013rare2012}  & SPIC2013 & T    & 0.7700 & 0.6729 & 1.1513    & 0.4220  & 1.0090 & 0.4572  \\
\small AIM \cite{bruce2007attention}  & JoV2007 & T     & 0.7619 & 0.6647 & 0.8824    & 0.3419  & 1.2476 & 0.4096  \\
\small CAS \cite{goferman2011context}  & TPAMI2011 & T     & 0.7581 & 0.6402 & 1.0186    & 0.3848  & 1.0723 & 0.4319  \\
% \small DVACLI \cite{hou2008dynamic}  & NIPS2008 & T     & 0.7548 & 0.6584 & 1.0142    & 0.3762  & 1.1136 & 0.4518  \\
% \small IS \cite{hou2011image}  & TPAMI2011 & T     & 0.7461 & 0.6610 & 0.9907    & 0.3709  & 1.0897 & 0.4278  \\
% \small QSS \cite{schauerte2012quaternion}  & ECCV2012 & T     & 0.7233 & 0.6679 & 0.9116    & 0.3300  & 1.1431 & 0.4208  \\
% \small SSR \cite{seo2009nonparametric}  & CVPR2009 & T     & 0.7064 & 0.6482 & 0.8110    & 0.2999  & 1.5255 & 0.4124  \\
% \small SUN \cite{zhang2008sun}  & JoV2008 & T     & 0.6939 & 0.6260 & 0.7620    & 0.2770  & 1.2815 & 0.4124  \\
\small Itti98 \cite{itti1998model}  & TPAMI1998 & T     & 0.5434 & 0.5357 & 0.4081    & 0.1307  & 1.4964 & 0.3378  \\
\hline
\small CASNetII \cite{fan2018emotional} & CVPR2018   & D   & 0.8552 & 0.7398 & 1.9859    & 0.7054  & 0.5857 & 0.5806  \\
\small SAM-Res \cite{cornia2018predicting} & TIP2018  & D   & 0.8526 & 0.7396 & 2.0628    & 0.6897  & 1.1710 & 0.6122  \\
\small SalGAN \cite{pan2017salgan}  & CVPRW2017  & D    & 0.8498 & 0.7354 & 1.8620    & 0.6740  & 0.7574 & 0.5932  \\
\small SAM-VGG \cite{cornia2018predicting} & TIP2018  & D  & 0.8473 & 0.7305 & 1.9552    & 0.6630  & 1.2746 & 0.5986  \\
\small DVA \cite{wang2017deep}  & TIP2018 & D  & 0.8430 & 0.7257 & 1.9305    & 0.6631  & 0.6293 & 0.5848  \\
\small DeepGazeI \cite{kummerer2014deep} & ICLR2015   & D  & 0.8427 & 0.7232 & 1.7234    & 0.6144  & 0.6678 & 0.5717  \\
\small MLNet \cite{cornia2016deep} & ICPR2016   & D   & 0.8368 & 0.7399 & 1.9748    & 0.6633  & 0.8006 & 0.5819  \\
\small ICF \cite{kummerer2017understanding} & ICCV2017   & D   & 0.8330 & 0.6957 & 1.6134    & 0.5876  & 0.7084 & 0.5576  \\
\small eDN \cite{vig2014large} & CVPR2014   & D  & 0.8171 & 0.6180 & 1.1399    & 0.4518  & 1.1369 & 0.4112  \\
\small SALICON \cite{huang2015salicon} & ICCV2015  & D    & 0.8171 & 0.6180 & 1.1399    & 0.4518  & 1.1369 & 0.4112  \\
\small MSI-Net \cite{kroner2020contextual} & NN2020  & D    & 0.8738 & 0.7787 & 2.3053    & 0.7790  & 0.4232 & 0.6704  \\
\small DeepGazeII \cite{kummerer2017understanding}  & ICCV2017 & D   & 0.8733 & 0.7759 & 2.3371    & 0.7703  & 0.4239 & 0.6636  \\
\small GazeGAN \cite{che2019gaze}  & TIP2019  & D    & 0.8607 & 0.7316 & 2.2118    & 0.7579  & 1.3390 & 0.6491  \\
\small EMLNet \cite{jia2020eml} & IVC2020  & D   & 0.8762 & 0.7469 & \textcolor{blue}{2.4876}    & \textcolor{green}{0.7893}  & 0.8439 & \textcolor{green}{0.6756}  \\
\small UNISAL \cite{droste2020unified} & ECCV2020  & D    & \textcolor{blue}{0.8772} & 0.7840 & 2.3689    & 0.7851  & \textcolor{blue}{0.4149} & 0.6746  \\
\small DeepGazeIIE \cite{linardos2021deepgaze} & ECCV2021  & D   & \textcolor{red}{0.8829} & \textcolor{red}{0.7942} & \textcolor{red}{2.5265}    & \textcolor{red}{0.8242}  & \textcolor{red}{0.3474} & \textcolor{red}{0.6993}  \\
\hline
% \small \textbf{SalFBNet-Res18 (our)}  & -  & D   & 0.8675 & 0.7685 & 2.2837    & 0.7691  & 0.4863 & 0.6601  \\
\small \textbf{SalFBNet-Res18 (Ours)}  & -  & \textbf{D}   & \textbf{\textcolor{green}{0.8769}} & \textbf{\textcolor{blue}{0.7858}} & \textbf{\textcolor{green}{2.4702}}    & \textbf{\textcolor{blue}{0.8141}}  & \textbf{\textcolor{green}{0.4151}} & \textbf{\textcolor{blue}{0.6933}}  \\
\hline
\multicolumn{9}{l}{~\small Note that all the metric values are collected from the MIT300 \cite{judd2012benchmark} leaderboard.}  \\
\end{tabular}
}
\label{tab:mit300_testing}
\end{table*}

\begin{table*}[!t]
\caption{
Quantitative comparison with 11 existing methods on testing sets of DUT-OMRON \cite{yang2013saliency}, PASCAL-S \cite{li2014secrets}, and TORONTO \cite{bruce2007attention}. The best three results are respectively shown in \textcolor{red}{red}, \textcolor{blue}{blue}, and \textcolor{green}{green} color.
}
\label{tab:dut_pas_tor_testing}
\resizebox{\textwidth}{!}{%
\begin{tabular}{@{}l|l|cccccc|cccccc|cccccc@{}}
\toprule
\multirow{2}{*}{} & \multirow{2}{*}{Method}  & \multicolumn{6}{c|}{DUT-OMRON \cite{yang2013saliency}} & \multicolumn{6}{c|}{PASCAL-S \cite{li2014secrets}} & \multicolumn{6}{c}{TORONTO \cite{bruce2007attention}} \\ 
\cmidrule(l){3-20}
 &  & AUC-J  & AUC-B  & sAUC  & CC  & NSS  & SIM & AUC-J  & AUC-B  & sAUC  & CC  & NSS & SIM  & AUC-J  & AUC-B  & sAUC  & CC  & NSS  & SIM  \\ \midrule
\multirow{4}{*}{\rotatebox[origin=c]{90}{Traditional}} 
 & CAS~\cite{goferman2011context}  & 0.8000 & 0.7900 & 0.7300 & 0.4000 & 1.4700 & 0.3700 & 0.7800 & 0.7500 & 0.6700 & 0.3600 & 1.1200 & 0.3400 & 0.7800 & 0.7800 & 0.6900 & 0.4500 & 1.2700 & 0.4400 \\
 & AIM~\cite{bruce2007attention} & 0.7700 & 0.7500 & 0.6900 & 0.3000 & 1.0500 & 0.3200 & 0.7700 & 0.7500 & 0.6500 & 0.3200 & 0.9700 & 0.3000 & 0.7600 & 0.7500 & 0.6700 & 0.3000 & 0.8400 & 0.3600 \\
 & GBVS~\cite{harel2006graph}  & 0.8700 & 0.8500 & 0.8100 & 0.5300 & 1.7100 & 0.4300 & 0.8400 & 0.8200 & 0.6500 & 0.4500 & 1.3600 & 0.3600 & 0.8300 & 0.8300 & 0.6400 & 0.5700 & 1.5200 & 0.4900 \\
 & Itti98~\cite{itti1998model}  & 0.8300 & 0.8300 & 0.7800 & 0.4600 & 1.5400 & 0.3900 & 0.8200 & 0.8000 & 0.6400 & 0.4200 & 1.3000 & 0.3600 & 0.8000 & 0.8000 & 0.6500 & 0.4800 & 1.3000 & 0.4500 \\
 \midrule
\multirow{7}{*}{\rotatebox[origin=c]{90}{Deep-Learning-based}}
& DVA~\cite{wang2017deep} & 0.9100 & 0.8600 & \textcolor{red}{0.8400} & \textcolor{green}{0.6700} & 3.0900 & 0.3900 & 0.8900 & \textcolor{blue}{0.8600} & \textcolor{red}{0.7600} & \textcolor{green}{0.7200} & 2.1200 & \textcolor{blue}{0.5800} & 0.8600 & \textcolor{red}{0.8600} & \textcolor{red}{0.7600} & 0.7200 & 2.1200 & 0.5800 \\
& *MSI-Net~\cite{kroner2020contextual} & \textcolor{green}{0.9149} & \textcolor{blue}{0.8947} & \textcolor{blue}{0.8187} & 0.6613 & 2.9350 & 0.4920 & 0.8931 & 0.8541 & \textcolor{blue}{0.7532} & 0.6603 & 2.2641 & 0.5264 & 0.8702 & 0.8263 & \textcolor{green}{0.7249} & 0.7419 & 2.1187 & 0.6253 \\
& *DeepGazeIIE~\cite{kummerer2017understanding} & 0.9044 & 0.8722 & 0.7647 & 0.6184 & 2.4690 & 0.4832 & \textcolor{red}{0.9125} & \textcolor{red}{0.8672} & 0.7338 & \textcolor{red}{0.7676} & \textcolor{blue}{2.5401} & \textcolor{red}{0.6136} & \textcolor{blue}{0.8805} & 0.8184 & 0.6964 & \textcolor{blue}{0.8034} & \textcolor{green}{2.3019} & \textcolor{red}{0.6659} \\
& *EMLNet~\cite{jia2020eml} & 0.9121 & 0.8628 & 0.8021 & 0.6620 & \textcolor{green}{3.1418} & \textcolor{blue}{0.5281} & 0.8897 & 0.8249 & 0.7406 & 0.6636 & 2.3581 & 0.5464 & 0.8546 & 0.7818 & 0.6990 & 0.7291 & 2.1901 & 0.6142 \\
& *CASNetII~\cite{fan2018emotional} & 0.9003 & 0.8832 & 0.8024 & 0.6012 & 2.6105 & 0.4356 & 0.8856 & 0.8518 & \textcolor{blue}{0.7473} & 0.6356 & 2.1741 & 0.4915 & 0.8629 & 0.8328 & 0.7175 & 0.6968 & 1.9447 & 0.5781 \\
& *UNISAL~\cite{droste2020unified} & 0.8913 & 0.8413 & 0.7779 & 0.5577 & 2.7776 & 0.4235 & 0.8766 & 0.7997 & 0.7212 & 0.5763 & 2.0919 & 0.4766 & 0.8692 & 0.7785 & 0.7067 & 0.6965 & 2.2050 & 0.5953 \\
& *FBNet~\cite{ding2021fbnet} & 0.9050 & \textcolor{green}{0.8928} & 0.8022 & 0.5848 & 2.5051 & 0.3959 & 0.8889 & \textcolor{green}{0.8685} & 0.7498 & 0.6121 & 2.0568 & 0.4540 & 0.8697 & \textcolor{blue}{0.8467} & 0.7094 & 0.7141 & 1.9707 & 0.5759 \\
% \cmidrule(l){2-20}
 \midrule
\multirow{4}{*}{\rotatebox[origin=c]{90}{\textbf{Ours}}}
& \bf{SalFBNet-Res18}  & \textbf{0.9118} & \textbf{0.8403} & \textbf{0.7740} & \textbf{0.6339} & \textbf{3.0879} & \textbf{0.4929} & \textbf{0.9043} & \textbf{0.7964} & \textbf{0.7155} & \textbf{0.6784} & \textbf{2.4997} & \textbf{0.5611} & \textbf{\textcolor{green}{0.8794}} & \textbf{0.7887} & \textbf{0.6942} & \textbf{0.7619} & \textbf{\textcolor{blue}{2.3314}} & \textbf{\textcolor{green}{0.6382}}\\
& \bf{SalFBNet-Res18Fixed} & \textbf{0.9144} & \textbf{\textcolor{red}{0.8949}} & \textbf{\textcolor{green}{0.8179}} & \textbf{0.6417} & \textbf{2.8716} & \textbf{0.4622} & \textbf{0.8943} & \textbf{0.8581} & \textbf{\textcolor{green}{0.7523}} & \textbf{0.6518} & \textbf{2.2429} & \textbf{0.5081} & \textbf{0.8756} & \textbf{\textcolor{green}{0.8354}} & \textbf{\textcolor{blue}{0.7264}} & \textbf{0.7475} & \textbf{2.1340} & \textbf{0.6200}\\ 
& \bf{$\dagger$SalFBNet-Res18}  & \textbf{\textcolor{red}{0.9269}} & \textbf{0.8856} & \textbf{0.8145} & \textbf{\textcolor{red}{0.7275}} & \textbf{\textcolor{red}{3.3439}} & \textbf{\textcolor{red}{0.5387}} & \textbf{\textcolor{blue}{0.9072}} & \textbf{0.8395} & \textbf{0.7434} & \textbf{\textcolor{blue}{0.7336}} & \textbf{\textcolor{red}{2.5808}} & \textbf{\textcolor{green}{0.5730}} & \textbf{\textcolor{red}{0.8824}} & \textbf{0.8114} & \textbf{0.7015} & \textbf{\textcolor{red}{0.8103}} & \textbf{\textcolor{red}{2.3647}} & \textbf{\textcolor{blue}{0.6576}}\\
& \bf{$\dagger$SalFBNet-Res18Fixed} & \textbf{\textcolor{blue}{0.9250}} & \textbf{0.8826} & \textbf{0.8092} & \textbf{\textcolor{blue}{0.7148}} & \textbf{\textcolor{blue}{3.2755}} & \textbf{\textcolor{green}{0.5016}} & \textbf{\textcolor{green}{0.9060}} & \textbf{0.8367} & \textbf{0.7382} & \textbf{0.7186} & \textbf{\textcolor{green}{2.5271}} & \textbf{0.5429} & \textbf{0.8782} & \textbf{0.8099} & \textbf{0.6948} & \textbf{\textcolor{green}{0.7826}} & \textbf{2.2666} & \textbf{0.6288}\\ 
\bottomrule
\multicolumn{20}{l}{~~~~~*~\small denotes the results are generated by their source code with default settings. Others are collected from the study~\cite{wang2017deep}.}  \\
\multicolumn{20}{l}{~~~~~$\dagger$~\small denotes the model is finetuned on the mixture training set of Pseudo-Saliency (900 samples) and MIT1003 (900 samples).}  \\
\end{tabular}%
}
\end{table*}

For the evaluation of MIT300 \cite{judd2012benchmark} dataset, we first use MIT1003 \cite{judd2009learning} to fine-tune our feedback model, then submit our results to MIT300 \cite{judd2012benchmark} benchmark.
In this work, we divide MIT1003 \cite{judd2009learning} into training set and validation set for fine-tuning, as shown in Table \ref{tab:comp_datasets}. We show the quantitative experimental results in Table \ref{tab:mit300_testing} on MIT300 testing set, including 16 existing deep-learning-based and 8 traditional-based approaches over 6 commonly used evaluation metrics. Note that all metric values of Table \ref{tab:mit300_testing} are collected from the benchmark leaderboard\footnote{https://saliency.tuebingen.ai/results.html} of MIT300. We can observe that the feedback model (SalFBNet-Res18) surpasses most existing saliency methods by a large margin. For example, the CC value and SIM value of our model achieve 0.8141 and 0.6933, respectively. For these two metrics, only our feedback model and DeepGazeIIE \cite{linardos2021deepgaze} obtain the performance of CC$\geq$0.81 and SIM$\geq$0.69. In addition, most metric values of our feedback model are ranked in the top three, and the performance gap between proposed model and the best model (DeepGazeIIE \cite{linardos2021deepgaze}) is very small (\emph{e.g.} sAUC: 0.84\%, CC: 1.01\%, SIM: 0.6\%).

\begin{figure}[!t]
    \centering
    \includegraphics[width=\textwidth]{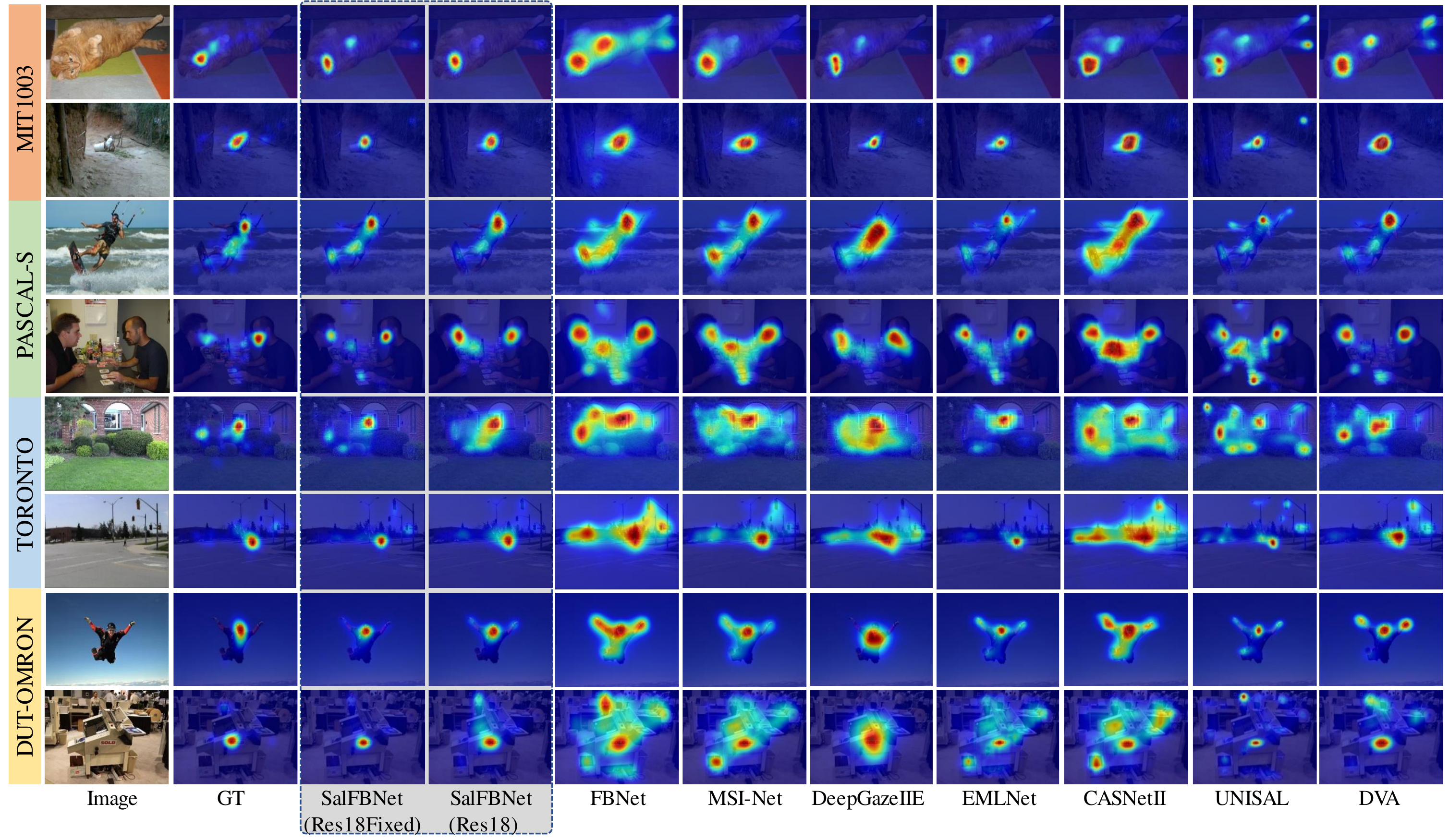}
    \caption{Visualization comparison on different datasets, including MIT1003 \cite{judd2009learning}, PASCAL-S \cite{li2014secrets}, TORONTO \cite{bruce2007attention}, and DUT-OMRON \cite{yang2013saliency}. From left to right: Image, ground-truth (GT), SalFBNet-Res18Fixed128 (Ours), SalFBNet-Res18 (Ours), FBNet \cite{ding2021fbnet}, MSI-Net \cite{kroner2020contextual}, DeepGazeIIE \cite{linardos2021deepgaze}, EMLNet \cite{jia2020eml}, CASNetII \cite{fan2018emotional},   UNISAL \cite{droste2020unified},  and DVA \cite{wang2017deep}.}
\label{fig:vis_four_datasets}
\end{figure}

Moreover, we compare our feedback models with 11 existing saliency methods on other three popular benchmarks, including DUT-OMRON \cite{yang2013saliency}, PASCAL-S \cite{li2014secrets}, and TORONTO \cite{bruce2007attention}. These existing models contain 4 conventional-based methods and 7 deep-learning-based models. In Table \ref{tab:dut_pas_tor_testing}, we show the quantitative comparison results. We first collect the available metric values from the study~\cite{wang2017deep}. For other models, we use their released code for the inference of saliency prediction with the default settings. After that, we evaluate all models with the same evaluation implementation (see Section \ref{sec_experiment}). In this experiment, we use a mixture training set with Pseudo-Saliency (900 samples) and MIT1003 (900 samples) for fine-tuning. From Table \ref{tab:dut_pas_tor_testing}, we can find that our feedback models achieve competitive quantitative results on these three datasets. Our two feedback models are also ranked on the top three in most metrics. Besides, our previous extremely lightweight FBNet \cite{ding2021fbnet} achieves the second best performance in terms of AUC-B metric on TORONTO \cite{bruce2007attention}. Furthermore, the feedback models finetuned with the mixture training set gain additional improvements and achieve the best performance. For instance, SalFBNet-Res18 achieves the best NSS performance on all three datasets after fine-tuning on the mixture training set. These results show that the performance can be further improved with the pseudo-saliency data. Finally, we visualize some challenging scenes and saliency predictions of different models in Figure \ref{fig:vis_four_datasets}. We can observe that our SalFBNet-Res18 and SalFBNet-Res18Fixed models can obtain better visual results in complex scenes compared with the state-of-the-art models.

\subsection{Ablation Studies}\label{sec_ablation}
In this section, we conduct extensive ablation studies to better understand the influence of various experimental settings in the proposed architecture.

% \input{FBNet_Tab_filtersize_comp}

% \input{FBNet_Tab_fbfw_comp}

% \input{FBNet_Fig_fwfb_visualize}

% \subsubsection{Ablation of fixed filter size}\label{sec_ablation_fixed_filter}
% First, we explore optimal filter size of backbone to build a lightweight yet efficient feedback model based on our previous work \cite{ding2021fbnet}. In the study \cite{ding2021fbnet}, we use a fixed filter size of 64 for all convolutional layers of the feature extractor. This not only contributes to the compactness of the feedback model, but also makes it easy to bridge the feed-forward and feed-back features. In this work, we first train extended models with various filter sizes (\emph{e.g.} 96, 128, \emph{etc.}), then we use $k$-folds cross validation on SALICON \cite{jiang2015salicon} dataset. In Table \ref{tab:filtersize_comp}, we show the averaged performance of $k$-folds cross validation. Here, we select $k$=3 in our experiment. From Table \ref{tab:filtersize_comp}, we can observe that as the number of convolution kernels increases, the performance of the model gets better and better. However, when the number of convolution kernels exceeds 128, the performance of the model no longer improves and even declines a little. Therefore, we choose 128 for the model with fixed filter size in our experiment.

% \input{FBNet_Tab_pseudosaliency_comp_backbones}

\begin{table}[!t]
\centering \caption{Quantitative performance of pseudo-annotators on SALICON \cite{jiang2015salicon} validation set. \textbf{Pseudo-Saliency} means the result of aggregated saliency prediction.}
\resizebox{0.86\textwidth}{!}{
\begin{tabular}{  c | c  c  c  c  c  c  c  }
\hline
% \small SALICON \cite{jiang2015salicon}  &  \multicolumn{7}{c}{\textbf{Validation Set}}   \\
\hline
\small Metric              & \small AUC-J $\uparrow$ &  \small AUC-B $\uparrow$   &  \small  sAUC $\uparrow$   & \small  CC $\uparrow$ & \small  NSS $\uparrow$ & \small  KLdiv $\downarrow$ & \small  SIM $\uparrow$ \\
\hline\hline
\small MSINet  \cite{kroner2020contextual}  & 0.8678     & 0.8332  & 0.7261 & 0.8886 & 1.9040    & 0.2752  & 0.7816 \\
\small DeepGaze  \cite{linardos2021deepgaze}              & 0.8559 & 0.7604     & 0.6572  & 0.7931 & 1.7258 & 0.5482 & 0.6927   \\
\small EMLNet  \cite{jia2020eml} & 0.8665 & 0.7974     & 0.7129  & 0.8789 & 1.9924 & 0.7363 & 0.7691  \\
\small CASNet   \cite{fan2018emotional}  & 0.8569 & 0.8283     & 0.7148  & 0.8416 & 1.7814 & 0.3523 & 0.7194  \\
\small UNISAL \cite{droste2020unified} & 0.8490 & 0.7656     & 0.6935  & 0.7710 & 1.8401 & 0.7479 & 0.6827  \\
\hline
\textbf{\small Pseudo-Saliency}  & \textbf{0.8706} & \textbf{0.8315}  & \textbf{0.7236}   & \textbf{0.8961} & \textbf{1.9591} & \textbf{0.2555} & \textbf{0.7784} \\
\hline
\end{tabular}
}
\label{tab:pseudo_comp}
\end{table}

\subsubsection{Analysis of Transferred Pseudo-Saliency}\label{sec_results}
\begin{figure*}
    \centering
    \includegraphics[width=\textwidth]{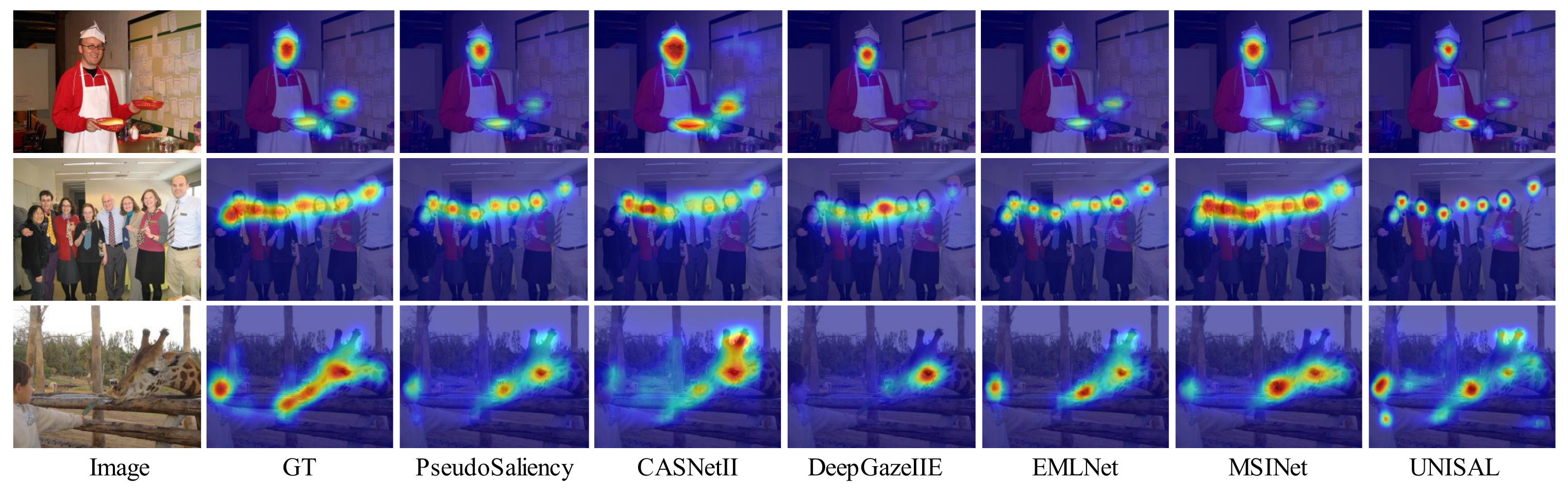}
    \caption{Visualization results of different model-annotators. The results are collected from annotators including MSINet \cite{kroner2020contextual}, DeepGazeIIE \cite{linardos2021deepgaze}, EMLNet \cite{jia2020eml}, CASNetII \cite{fan2018emotional}, and UNISAL \cite{droste2020unified}. PseudoSaliency means the aggregated pseudo-saliency in this work. }
\label{fig:pseudo-visual}
\end{figure*}

We first conduct experiments to evaluate the effectiveness of the pseudo-saliency knowledge transferred from \emph{model-annotators}. We compare quantitative performance between Pseudo-Saliency and the five model-annotators on SALICON \cite{jiang2015salicon} validation set as shown in Table \ref{tab:pseudo_comp}. In addition, we show several saliency prediction samples of different annotators in Figure \ref{fig:pseudo-visual}. More detailed analysis on the selected saliency models and aggregated pseudo saliency results can be found in Figure 1 and Table 3 of Supplementary file. From these experiments, we verify the following observations:

i) All pseudo-annotators can predict roughly correct distribution by visually and quantitatively comparing with the ground-truth, but some predicted gazes are shifted and incomplete. For instance, the prediction of CASNetII \cite{fan2018emotional} is shifted to the animal head in the third row of Figure \ref{fig:pseudo-visual}.
%while DeepGazeIIE \cite{linardos2021deepgaze} predicts incomplete results, as shown in the first and third rows of Figure \ref{fig:pseudo-visual}.

ii) These methods are able to predict similar saliency distributions. However, there are biases from different models due to various model settings and training strategies. 
%For example, UNISAL \cite{droste2020unified} considers the saliency of images and videos in a unified model. 
Therefore, we aggregate predicted distributions of these models to alleviate deviations from different model-annotators.

iii) The aggregated pseudo-saliency not only combines the advantages of each annotator, but also suppresses some potential defective predictions. Quantitatively, the metric values of PseudoSaliency are always close to the best model among these model-annotators, and even achieve superior performance, such as AUC-J and KLdiv in Table \ref{tab:pseudo_comp}.

\subsubsection{Influences of backbones and training settings}\label{sec_ablation_backbones_pretraining}
We extend the feature extractor of the feedback model with different popular backbones to further demonstrate the flexibility, stability and consistency of the proposed framework, including ResNet \cite{he2016deep}, DenseNet \cite{huang2017densely}. We also modify a feedback model, namely Res18Fixed128, to compare with our previous work \cite{ding2021fbnet}. The backbone of Res18Fixed128 is ResNet18 \cite{he2016deep} with fixed filter size of 128 in all convolution layers. In this work, we experimentally select 128 for the model with fixed filter size. The details and results of filter size selection experiments are given in Table 1 and Table 2 of Supplementary file. For training, we use different initialization settings to verify the pre-trained weights on ImageNet \cite{krizhevsky2012imagenet}. 
%For fair comparison, we train all models on Pseudo-Saliency dataset and evaluate their performance on SALICON \cite{jiang2015salicon} validation set. 
%Note that we use Mean Square Error (MSE) loss to replace sFNE loss of Equation (\ref{equ:combined_loss}) for the training on Pseudo-Saliency dataset because there is no fixation map. 
%In this work, we select three feedback models to verify the fine-tuning performance of the hybrid loss. Their backbones are Densenet121, ResNet18, and ResNet50, respectively. 
We first pre-train these feedback models on Pseudo-Saliency dataset without using any human annotated ground truth of public datasets. Extensive results with pre-trained performance of our proposed framework using various backbone models on SALICON \cite{jiang2015salicon} validation set are shown and compared in Table 4 of Supplementary file. We then finetune them on SALICON \cite{jiang2015salicon} with proposed loss in Equation (\ref{equ:combined_loss}). 
%In this experiment, we also use different initialization strategies to validate the performance of fine-tuning. 
%We show comparison results in Table \ref{tab:finetune_comp_backbones}. 
%From Table \ref{tab:finetune_comp_backbones}, we can clearly observe that all models gain additional improvements on most metrics after fine-tuning with the hybrid loss. These experiments indicate that proposed fixation-based sFNE loss can further improve the performance for saliency prediction. 
From Table \ref{tab:finetune_comp_backbones}, we can draw the following conclusions:

i) There is a slight influence between feedback models initialized by ImageNet \cite{krizhevsky2012imagenet} pre-trained weights and random initialization. In particular, the performance of some randomly initialized models even surpasses that of initialized by ImageNet \cite{krizhevsky2012imagenet} pre-trained weights.

\begin{table*}[!t]
\centering \caption{Comparison of finetuning performance of feedback models with different backbones and initialization strategies.}
\resizebox{\textwidth}{!}{
\begin{tabular}{  c | c | c | c | c  c  c  c  c  c  c }
\hline
% \small Dataset  &  \multicolumn{8}{c}{SALICON \cite{jiang2015salicon} validation set}   \\
% \hline
\small Backbone   &  Init.  & Train & Eval.  & \small AUC-J $\uparrow$ &  \small AUC-B $\uparrow$   &  \small  sAUC $\uparrow$   & \small  CC $\uparrow$ & \small  NSS $\uparrow$ & \small  KLdiv $\downarrow$ & \small  SIM $\uparrow$  \\
\hline\hline

\small Densenet121 & ImageNet   & SALICON & SALICON  & 0.8395  & 0.8290 & 0.6914 & 0.7565    & 1.5184  & 0.4620 & 0.6608   \\
\small Res18       & ImageNet  & SALICON & SALICON   & 0.8379  & 0.8221 & 0.6838 & 0.7460    & 1.5079  & 0.4984 & 0.6597     \\
\small ResNet50  & ImageNet    & SALICON & SALICON   & 0.8509  & 0.8310 & 0.7058 & 0.8087    & 1.6853  & 0.4026 & 0.6835   \\
\hline
\small Densenet121  & Random    & SALICON & SALICON  & 0.8392  & 0.8289 & 0.6882 & 0.7547    & 1.5121  & 0.4576 & 0.6595  \\
\small ResNet18       & Random   & SALICON & SALICON   & 0.8340  & 0.8207 & 0.6782 & 0.7315    & 1.4643  & 0.5870 & 0.6537  \\
\small ResNet50  & Random   & SALICON & SALICON   & 0.8538  & 0.8309 & 0.7110 & 0.8139    & 1.7054  & 0.4459 & 0.7098 \\
\small Res18Fixed128  & Random     & SALICON & SALICON  & 0.8551  & 0.8341 & 0.7120 & 0.8234    & 1.7331  & 0.3996 & 0.6744  \\
\hline

\small Densenet121 & ImageNet + PseudoSal.   & SALICON & SALICON  & 0.8670  & 0.8493 & 0.7314 & 0.8804    & 1.8584  & 0.2649 & 0.7559   \\
\small ResNet18       & ImageNet + PseudoSal.   & SALICON & SALICON   & 0.8680  & 0.8467 & 0.7344 & 0.8851    & 1.8815  & 0.3014 & 0.7673     \\
\small ResNet50  & ImageNet + PseudoSal.    & SALICON & SALICON   & 0.8676  & 0.8501 & 0.7338 & 0.8810    & 1.8616  & 0.2768 & 0.7480   \\
\hline
\small Densenet121  & Random + PseudoSal.    & SALICON & SALICON  & 0.8660  & 0.8466 & 0.7297 & 0.8760    & 1.8542  & 0.2850 & 0.7556  \\
\small ResNet18       & Random + PseudoSal.   & SALICON & SALICON   & 0.8682  & 0.8474 & 0.7337 & 0.8860    & 1.8778  & 0.2700 & 0.7675  \\
\small ResNet50  & Random + PseudoSal.   & SALICON & SALICON   & 0.8673  & 0.8490 & 0.7347 & 0.8812    & 1.8620  & 0.2692 & 0.7480 \\
\small Res18Fixed128  & Random + PseudoSal.    & SALICON & SALICON  & 0.8696  & 0.8421 & 0.7331 & 0.8917    & 1.9182  & 0.2361 & 0.7684  \\
\hline
\end{tabular}
}
\label{tab:finetune_comp_backbones}
\end{table*}

ii) Feedback models with fixed filter size achieve competitive performance despite their small number of parameters. Some metric values of these lightweight models even outperform those with popular backbones. These results reveal that the feedback model with fixed filter size can also capture plenty of informative representations for saliency prediction.

% iii) All models have learned visual representations and generalization capabilities from pseudo-saliency data, which indicates that the human-free annotations of Pseudo-Saliency dataset contain the saliency knowledge transferred from model-annotators. Note that our Pseudo-Saliency data used in the training is a large-scale \emph{model-annotated} dataset without human-annotations.

% iii) All models gain additional improvements on most metrics after fine-tuning with the hybrid loss. These experiments indicate that proposed fixation-based sFNE loss can further improve the performance for saliency prediction. 

The above findings further validate conclusions of the study \cite{cheng2021highly} that, it is not necessary to adopt the pre-trained weights of ImageNet \cite{krizhevsky2012imagenet} and a backbone with a large number of parameters for saliency detection. 
%Therefore, based on our empirical evaluation, we argue that this is due to the difference between the image recognition task and the pixel-wise saliency prediction task.

\subsubsection{Ablation study of hybrid loss}\label{sec_ablation_hybrid_loss}
Moreover, we experimentally demonstrate the effectiveness of the proposed hybrid loss in this work. Specifically, we select the feedback model with ResNet18 backbone to investigate the influence of the combined loss in Equation (\ref{equ:combined_loss}) for saliency prediction. 
%After this model is pre-trained on Pseudo-Saliency dataset, we use the loss of Equation (\ref{equ:combined_loss}) for fine-tuning. 
Here, we adopt the loss of the study \cite{droste2020unified} as a baseline. 
%In addition, we search the optimal hyper-parameters for the combined loss. 
From Table \ref{tab:combined_loss}, we can observe that the loss of the study \cite{droste2020unified} can achieve best NSS performance due to the combination of NSS loss (-NSS). However, it cannot obtain better results in terms of distribution-based metrics, such as CC and KLdiv. On the other hand, the 1-CC loss can compensate the performance degradation of saliency prediction to some extent. Also, the loss combined with the proposed sFNE significantly improves the performance of the feedback model in most metrics, as shown in Table \ref{tab:combined_loss}. These experimental results show the effectiveness of the proposed sFNE loss, since it considers not only the distribution at fixations, but also covers the normalized saliency at non-fixations. 

\begin{table*}[!t]
\centering \caption{Ablation studies of the hybrid loss.}
\resizebox{\textwidth}{!}{
\begin{tabular}{  c | c | c  c  c  c  c  c  c }
\hline
\small Combined Loss     &  Stage    & \small AUC-J $\uparrow$ &  \small AUC-B $\uparrow$   &  \small  sAUC $\uparrow$   & \small  CC $\uparrow$ & \small  NSS $\uparrow$ & \small  KLdiv $\downarrow$ & \small  SIM $\uparrow$ \\
\hline\hline
\small 1*KLD+0.1*(1-CC)+0.025*MSE  & Pre-training     & 0.8641  & 0.8400 & 0.7176 & 0.8665    & 1.8483  & 0.3122 & 0.7209  \\
\small 1*KLD+0.1*(1-CC)+0.025*MSE  & Finetuning    & 0.8667  & 0.8410 & 0.7269 & 0.8828    & 1.8746  & 0.2491 & 0.7594  \\
\hline
\small 1*KLD+0.1*(-CC)+0.1*(-NSS) \cite{droste2020unified} & Finetuning   & 0.8667  & 0.8077 & 0.7116 & 0.8657    & 1.9303  & 0.4157 & 0.7560  \\
\small 1*KLD+0.1*(1-CC)+0.1*(-NSS) & Finetuning     & 0.8644  & 0.8177 & 0.7149 & 0.8624    & 1.8904  & 0.3192 & 0.7500  \\
\small 1*KLD+0.1*(1-CC)+0.01*(-NSS) & Finetuning     & 0.8647  & 0.8342 & 0.7213 & 0.8712    & 1.8680  & 0.2735 & 0.7519  \\
\hline
% \small 1*KLD+0.1*(1-CC)+0.1*(-NSS)+0.1*sfe & Finetuning     & 0.8666  & \textbf{0.8497} & \textbf{0.7338} & 0.8706    & 1.8514  & 0.2773 & 0.7412  \\
\small \textbf{1*KLD+0.1*(1-CC)+0.025*sFNE} & \textbf{Finetuning}     & \textbf{0.8682}  & \textbf{0.8474} & \textbf{0.7337} & \textbf{0.8860}    & \textbf{1.8778}  & \textbf{0.2700} & \textbf{0.7675}  \\
\hline
\end{tabular}
}
\label{tab:combined_loss}
\end{table*}

\subsection{Efficiency Comparison}
\label{sec_efficiency}

In Table \ref{tab:comp_params}, we show efficiency comparison of different models. The values of model size and run-time of other works are taken from the literature \cite{droste2020unified}. 
%We use \emph{torchSummeryX}\footnote{https://github.com/sksq96/pytorch-summary} toolbox to calculate the number of parameters for PyTorch model implementations. 
From Table \ref{tab:comp_params}, we observe that our feedback models have fewer number of parameters and small model storage size. Also, it can achieve relatively high frame-per-second (\emph{i.e.} around 20 fps ) processing during run-time.

\begin{table}[!t]
\centering \caption{Efficiency comparison of saliency methods.}
\resizebox{0.86\textwidth}{!}{
\begin{tabular}{  c | c| c  c  c  c  }
\hline
\small Model & Implem. & \#Params (M) & \#Trainable (M) & Size (MB) &  Runtime (s)   \\
\hline\hline
\small ShallowNet \cite{pan2016shallow} & Theano  & - & - & 2500  & 0.100  \\
\small CASNetII \cite{fan2018emotional} & Tensorflow  & - & - & 1100.0  & 1.220  \\
\small SalGAN \cite{pan2017salgan} & Theano  & - & - & 130.0  & 0.02  \\
\small SALICON \cite{huang2015salicon} & Caffe  & - & - & 117.0  & 0.500  \\
\small DeepNet \cite{pan2016shallow} & Caffe  & - & - & 103.0  & 0.080 \\
\small DVA \cite{wang2017deep} & Caffe  & - & - & 96.0  & 0.100  \\
\small MSI-Net \cite{kroner2020contextual} & Tensorflow  & - & - & 95.2  & 0.282  \\
\small DeepGazeIIE \cite{kummerer2017understanding} & PyTorch & 75.88 & 3.06 & 401.0  & 5.943  \\
\small EMLNet \cite{jia2020eml} & PyTorch  & 47.30 & 47.30  & 180.2  & 0.023  \\
\hline
\small FBNet \cite{ding2021fbnet} & PyTorch  & 1.18 & 1.18  & 4.7  & 0.029  \\
\small SalFBNet-Res18 & PyTorch  & 17.26 & 17.26  & 67.9  & 0.049  \\
\small SalFBNet-Res18Fixed & PyTorch  & 5.97 & 5.97  & 23.4  & 0.047  \\
\hline
% \multicolumn{6}{l}{~\small We use the \emph{torchSummaryX} toolbox to calculate the number of total parameters and trainable parameters.}  \\
\end{tabular}
}
\label{tab:comp_params}
\end{table}

\section{Conclusion}
\label{sec:conclusion}
In this paper, we propose a novel feedback convolutional architecture to learn abundant contextual features for saliency prediction. The proposed feedback model bridges the pathways from high-level feature blocks to low-level layer with feedback convolutional connections. Besides, we propose a new fixation-based loss, namely Selective Fixation and Non-Fixation Error (sFNE), to facilitate the proposed feedback model to better learn distinguishable eye-fixation-based features. Furthermore, we propose a Pseudo-Saliency dataset to alleviate the problem of data deficiency in the field of image saliency detection. We experimentally show the effectiveness of the proposed sFNE loss and Pseudo-Saliency dataset. Extensive experimental evaluation on various benchmarks has promising results by giving performance better or on-par with the state-of-the-arts saliency prediction models. Additionally, we observe that it is not necessary to use the backbone with a large number of parameters and weights of the pre-trained models on ImageNet \cite{krizhevsky2012imagenet} for saliency prediction.

\section{ACKNOWLEDGMENT}
\label{sec:ACKNOWLEDGMENT}

%The authors are thankful to *** for their valuable comments to improve the quality of this paper. This work is in part supported by ***.

This paper is in part based on the results obtained from a project commissioned by the New Energy and Industrial Technology Development Organization (NEDO), Japan.

% \section*{References}
% \small
\bibliography{References}

\end{document}